\documentclass{article}
\pdfoutput=1

\usepackage{sty/arxiv}

\usepackage[utf8]{inputenc}
\usepackage[T1]{fontenc}
\usepackage{natbib}
\usepackage{hyperref}
\usepackage{url}
\usepackage{booktabs}
\usepackage{amsfonts}
\usepackage[pdftex]{graphicx}
\usepackage{amsmath}
\usepackage{amssymb}
\usepackage{bm}
\usepackage{subcaption}
\usepackage{algorithm}
\usepackage{algpseudocode}
\usepackage{multirow}
\usepackage{float}
\usepackage{comment}
\usepackage{wrapfig}
\usepackage{tikz}
\usetikzlibrary{positioning, fit, arrows.meta, calc}

\title{Deep Probabilistic Supervision \\ for Image Classification}

\author{
    Anton Adelöw\textsuperscript{1}, 
    Matteo Gamba\textsuperscript{2}, 
    \& Atsuto Maki\textsuperscript{1} \\
    \textsuperscript{1}Division of Robotics, Perception and Learning, KTH Royal Institute of Technology \\
    \textsuperscript{2}Computer Science Department, Brown University \\
}

\begin{document}
\maketitle

\begin{abstract}

Supervised training of deep neural networks for classification typically relies on hard targets, which promote overconfidence and can limit calibration, generalization, and robustness. Self-distillation methods aim to mitigate this by leveraging inter-class and sample-specific information present in the model’s own predictions, but often remain dependent on hard targets without explicitly modeling predictive uncertainty. With this in mind, we propose Deep Probabilistic Supervision (DPS), a principled learning framework constructing sample-specific target distributions via statistical inference on the model’s own predictions, remaining independent of hard targets after initialization. We show that DPS consistently yields higher test accuracy (e.g., +2.0\% for DenseNet-264 on ImageNet) and significantly lower Expected Calibration Error (ECE) (-40\% ResNet-50, CIFAR-100) than existing self-distillation methods. 
When combined with a contrastive loss, DPS achieves state-of-the-art robustness under label noise. The source code is available at \url{https://github.com/antonadelow/DPS}.

\end{abstract}

\section{Introduction}
\label{sec:intro}
Despite the widespread use of deep neural networks for classification tasks, the implications of using hard targets for loss computation have received relatively little attention. Intuitively, classes can exhibit varying degrees of similarity, and individual samples may enjoy different levels of resemblance to both their assigned and other classes. This nuanced, sample-specific information -- often referred to as dark knowledge \citep{hinton2014dark} -- is discarded when relying solely on hard targets. For instance, misclassifying an image of a dog as a teapot is inherently different from misclassifying it as a cat. Nevertheless, deep networks are typically trained with hard targets, which fail to capture inter-class relationships or sample-specific ambiguities, contributing to poor calibration \citep{guo2017calibration}.

\textit{Self-distillation}, an efficient variant of \textit{knowledge distillation} \citep{hinton2015distilling}, leverages dark knowledge by combining a model's own predictions with the hard targets during training. In contrast, conventional knowledge distillation provides soft targets to a model by leveraging another, often larger, network. Despite their benefits, existing self-distillation methods exhibit varying limitations. Some modify the network architecture with intermediate classifiers \citep{zhang2019your, zhang2022self}, increasing parameter count and computational cost. Others impose a consistency loss between current predictions and those from the last epoch or mini-batch \citep{kim2021self, shen2022self}. Yet, epoch-wise construction introduces variance under augmentations, while targets from the last mini-batch may be of limited effectiveness, as they originate from nearly identical model states.

\begin{figure*}[h!]
    \centering
    \resizebox{\textwidth}{!}{%

\begin{tikzpicture}[
    node distance=1cm and 2.5cm, 
    neuron/.style={circle, draw, minimum size=0.8cm, thick},
    input/.style={neuron, fill=cyan!30},
    hidden/.style={neuron, fill=orange!30},
    output/.style={neuron, fill=green!30},
    container/.style={draw, thick, rounded corners, inner sep=0.8cm}, 
    arrow/.style={-Latex, thick, line width=1.5pt}
]

\node[input] (i1) {};
\node[input, below=0.5cm of i1] (i2) {};
\node[input, below=0.5cm of i2] (i3) {};


\coordinate (h1-anchor) at ([xshift=2.5cm]i2.center); 
\node[hidden, above=1.5cm of h1-anchor] (h1-1) {};
\node[hidden, above=0.5cm of h1-anchor] (h1-2) {};
\node[hidden, below=0.5cm of h1-anchor] (h1-3) {};
\node[hidden, below=1.5cm of h1-anchor] (h1-4) {};

\coordinate (h2-anchor) at ([xshift=2.5cm]h1-anchor);
\node[hidden, above=1.5cm of h2-anchor] (h2-1) {};
\node[hidden, above=0.5cm of h2-anchor] (h2-2) {};
\node[hidden, below=0.5cm of h2-anchor] (h2-3) {};
\node[hidden, below=1.5cm of h2-anchor] (h2-4) {};

\coordinate (o-anchor) at ([xshift=2.5cm]h2-anchor);
\node[output] (o2) at (o-anchor) {};
\node[output, above=0.5cm of o2] (o1) {};
\node[output, below=0.5cm of o2] (o3) {};

\foreach \i in {1,2,3}
    \foreach \j in {1,2,3,4}
        \draw[black!50] (i\i) -- (h1-\j);

\foreach \i in {1,2,3,4}
    \foreach \j in {1,2,3,4}
        \draw[black!50] (h1-\i) -- (h2-\j);

\foreach \i in {1,2,3,4}
    \foreach \j in {1,2,3}
        \draw[black!50] (h2-\i) -- (o\j);


\node[container, fit=(i1) (i3) (o3), label={[font=\bfseries\Large]above:Neural Network}] (network-container) {};
\node[below=0.2cm of network-container.south] {\Large $f(\mathbf{x}_i)$};

\node[container, left=1.5cm of network-container, align=center, label={[font=\bfseries\Large]above:Input}] (input-container) {\Large Input \\ \Large Image};
\node[below=0.2cm of input-container.south] {\Large $\mathbf{x}_i$};

\draw[arrow] (input-container.east) -- (network-container.west);

\node[right=3.5cm of network-container] (dist-placement-anchor) {};
\begin{scope}[shift=(dist-placement-anchor)]
    \begin{scope} 
        \draw[fill=blue!50, thick] (-1.4, 1.2) rectangle (0.6, 1.6);
        \draw[fill=blue!50, thick] (-1.4, 0.5) rectangle (-0.6, 0.9);
        \draw[fill=blue!50, thick] (-1.4, -0.2) rectangle (-0.8, 0.2);
        \draw[fill=blue!50, thick] (-1.4, -0.9) rectangle (-1.0, -0.5);
        \draw[fill=blue!50, thick] (-1.4, -1.6) rectangle (-1.2, -1.2);
        
        \draw[-latex, thick] (-1.4, -1.7) -- (-1.4, 1.9); 
        \draw[-latex, thick] (-1.4, -1.7) -- (1.4, -1.7); 

        \node[container, fit={(-1.3, -1.7) (1.3, 1.7)}, label={[font=\bfseries\Large]above:Predicted Target}] (dist-container) {};
        
        \node[below=0.2cm of dist-container.south] {\Large $\hat{\mathbf{y}}_i^t$};
    \end{scope}
\end{scope}

\draw[arrow] (network-container.east) -- (dist-container.west);

\node (target-container-node) [right=3.5cm of dist-container] {};
\begin{scope}[shift=(target-container-node.center)]
    \draw[-latex, thick] (-1.4, -1.7) -- (-1.4, 1.9);
    \draw[-latex, thick] (-1.4, -1.7) -- (1.4, -1.7);
    
    \draw[fill=green!50, thick] (-1.4, 1.2) rectangle (0.6, 1.6);
    
    \node[container, fit={(-1.3, -1.7) (1.3, 1.7)}, label={[font=\bfseries\Large]above:Prior Target}] (target-container) {};

    \node[below=0.2cm of target-container.south] {\Large $p(\mathrm{z}_i)=\text{Cat}\left(K,\frac{\bm{\alpha}_i}{\sum_j \bm{\alpha}_{ij}\}}\right)$};
\end{scope}

\node (updated-target-node) [right=5cm of target-container] {};
\begin{scope}[shift=(updated-target-node.center)]
    \draw[-latex, thick] (-1.4, -1.7) -- (-1.4, 1.9);
    \draw[-latex, thick] (-1.4, -1.7) -- (1.4, -1.7);
    
    \draw[fill=green!50, thick] (-1.4, 1.2) rectangle (0.6, 1.6);
    
    \draw[fill=blue!50, thick] (-1.4, 0.5) rectangle (-1.0, 0.9);
    \draw[fill=blue!50, thick] (-1.4, -0.2) rectangle (-1.1, 0.2);
    \draw[fill=blue!50, thick] (-1.4, -0.9) rectangle (-1.2, -0.5);
    \draw[fill=blue!50, thick] (-1.4, -1.6) rectangle (-1.3, -1.2);

    \node[container, fit={(-1.3, -1.7) (1.3, 1.7)}, label={[font=\bfseries\Large]above:Posterior Target}] (updated-target-container) {};

    \node[below=0.2cm of updated-target-container.south] {\Large $p(\mathrm{z}_i|\mathbf{x}_i,\hat{\mathbf{y}}_i^t)=\text{Cat}\left(K,\frac{\hat{\mathbf{y}}_i^t+\gamma\bm{\alpha}_i}{1+\sum_j \gamma\bm{\alpha}_{ij}}\right)$};
\end{scope}


\path (dist-container.east) -- (target-container.west) node[midway, font=\Huge] {$\oplus$};

\path (target-container.east) -- (updated-target-container.west) node[midway, font=\Huge] {$\times \gamma \rightarrow$};

\end{tikzpicture}
}
\caption{\textbf{Bayesian update of a target.} At each epoch, the target distributions are updated using the model's own predictions. Here, $\oplus$ denotes the Bayesian update, and $\times$ indicates discounting of the previous posterior parameters by $\gamma \in [0,1]$. See Section \ref{sec:method} for details.}
    \label{fig:target-refinement}
\end{figure*}
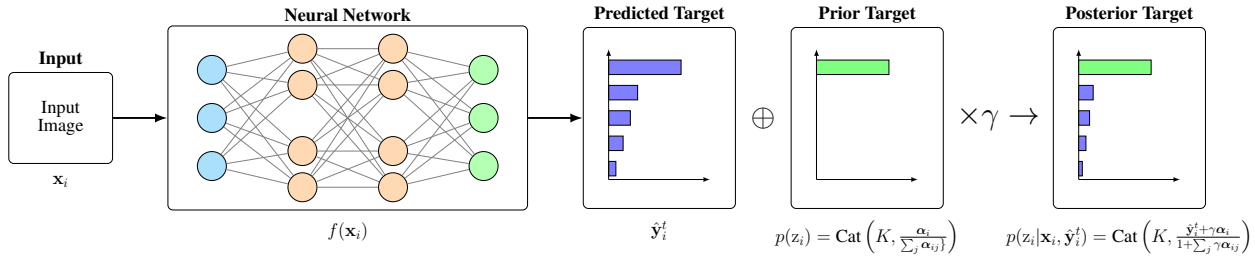

Fundamentally, existing self-distillation frameworks do not model predictive uncertainty explicitly, which may affect calibration negatively, while also constraining their predictions by incorporating a loss term derived from the hard targets (e.g. \citet{furlanello2018born, kim2021self, shen2022self}). This reliance interferes with the goal of learning more calibrated and nuanced predictions, leaving the model sensitive to label noise. In the overparameterized regime, sensitivity to label noise can exacerbate double descent \citep{nakkiran2021deep}, a phenomenon where the test error initially decreases, then increases, before decreasing again as model capacity or training time varies. 
In this work, we hypothesize that a self-distillation method decoupled from hard targets could provide smoother and more robust probability estimates, thereby mitigating these effects and providing more predictable training dynamics.

To address these limitations, we propose Deep Probabilistic Supervision (DPS), an efficient learning framework for leveraging dark knowledge by constructing nuanced, sample-specific target distributions in a single training run by modeling predictive uncertainty explicitly. DPS treats a model's own predictions as evidence for Bayesian inference, without relying on hard targets after initialization. A diagram depicting DPS is included in Figure \ref{fig:target-refinement}.

\noindent Our main contributions can be summarized as follows:
\begin{itemize}
    \item We propose Deep Probabilistic Supervision (DPS), a lightweight and principled framework for constructing nuanced target distributions during training that operates independently of hard targets after initialization.
    \item We demonstrate through extensive experiments that DPS consistently outperforms both existing self-distillation methods and conventional knowledge distillation in generalization and calibration. 
    \item We show that DPS provides robustness against data corruptions, perturbations, and label noise. Notably, DPS mitigates epoch-wise double descent under label noise and achieves state-of-the-art performance when combined with a contrastive loss.
    \item We provide the first, to our knowledge, theoretical formalization of dark knowledge and use it to show that DPS captures nuanced, sample-specific, information.
\end{itemize}

\section{Related work}
\label{sec:related_word}
The notion of Self-Distillation (SD) originates from Knowledge Distillation (KD) \citep{hinton2015distilling}, a technique introduced for model compression using a teacher–student framework. Later work demonstrated that identical (Born-Again) networks can be trained sequentially with KD to improve generalization \citep{furlanello2018born}, thereby giving rise to self-distillation. 

SD methods differ primarily in how they construct the teacher signal. One family of approaches relies on architectural structure, such as introducing auxiliary classifiers or branches to provide internal supervision \citep{zhang2019your, zhang2022self, lan2018knowledge}, while another leverages temporal knowledge transfer, using model snapshots from earlier epochs \citep{yang2019snapshot}, or moving averages of either past predictions (Temporal Ensembling, TE) \citep{laine2016temporal} or model parameters (Mean Teacher) \citep{tarvainen2017mean}. A third category enforces consistency, either between same-class samples \citep{yun2020regularizing} or consecutive predictions, including methods such as Progressive Self-Knowledge Distillation (PS-KD) \citep{kim2021self} and Self-Distillation from the Last mini-Batch (DLB) \citep{shen2022self} that leverage predictions from the previous epoch or mini-batch, respectively. Despite their differences, these strategies remain anchored to the original hard labels, which can be particularly problematic in settings where labels are noisy. 

Label noise remains an obstacle in deep learning, contributing to overfitting and undesirable training dynamics. Proposed solutions range from regularization techniques such as label smoothing \citep{szegedy2016rethinking} and Mixup \citep{zhang2017mixup}, to robust loss functions like Symmetric Cross Entropy \citep{wang2019symmetric}. More sophisticated methods include sample selection via optimal transport \cite{feng2023ot, chang2023csot}, label correction by interpolation between the prediction and label \citep{xu2025revisiting}, or complex multi-stage \citep{liu2023chimera} or multi-network training pipelines \citep{zhang2024bpt}, but often rely on heuristics rather than \textit{reformulating the learning objective to explicitly model uncertainty}. Probabilistic modeling provides a principled way to estimate uncertainty, improving interpretability.

Bayesian methods in deep learning have gained attention for their ability to model predictive uncertainty. Methods like Variational inference \citep{blundell2015weight}, Monte Carlo dropout \citep{gal2016dropout}, and deep ensembles \citep{lakshminarayanan2017simple} are used to quantify uncertainty, improve calibration or increase robustness to out-of-distribution data. Related work has also shown that KD itself can serve as a mechanism to transfer better calibration from teacher to student \citep{hebbalaguppe2024calibration}. 

Our proposed framework, Deep Probabilistic Supervision (DPS), is inspired by these domains. Like SD, DPS constructs softened target distributions, but models target construction probabilistically by leveraging model predictions as evidence. Distinct from other Bayesian methods that model distributions over weights or predictions, DPS models distributions over targets. DPS allows targets to evolve independently of the original labels, offering flexible self-supervision that inherently models uncertainty and increases robustness against label noise.

\section{Methodology}
\label{sec:method}
As discussed in Section \ref{sec:intro}, a network trained exclusively on hard targets is not incentivized to learn meaningful inter-class relationships or account for sample-specific ambiguities. To address this, and explicitly model uncertainty in target distributions, we frame the training process as Recursive Bayesian Estimation~\citep{singhal1988training}. Rather than viewing the network as a static generator of fixed predictions, we treat it as a sensor observing the external dataset under the stochastic dynamics of training, and aim to construct stable target distributions by recursively integrating predictions for improved generalization and reduced overconfidence. The proposed method, Deep Probabilistic Supervision (DPS), is summarized in Algorithm \ref{algo:refine_labels}.

\subsection{Notation}
Consider the supervised classification problem with a dataset \(
\mathcal{D} = \{(\mathbf{x}_i, \mathbf{y}^0_i)\}_{i=1}^{n}\) consisting of
samples \(\mathbf{x}_i \in \mathbb{R}^d\) and one-hot targets \(\mathbf{y}^0_i \in \Delta_k\). Let \(f: \mathbb{R}^d \to \mathbb{R}^k\) denote a neural network with parameters \(\bm\theta\) and a softmax activation as its last layer, let \(\ell\) be the loss function, and let \(\hat{\mathbf{y}}_i^t\) denote the prediction the model outputs for sample $i$ at epoch $t$.

\subsection{Deep Probabilistic Supervision} \label{sec:dps}
Training deep networks is a stochastic process where randomness enters through stochastic optimization (e.g., SGD), data augmentation, and regularization mechanisms. SGD dynamics approximate a posterior distribution over model parameters \citep{mandt2017stochastic}, while predictions under stochastic regularization have been suggested to represent draws from an implicit predictive distribution \citep{gal2016dropout}. Consequently, we consider the model’s prediction of a sample $\mathbf{x}_i$ as a random sample from a predictive distribution. 

From this perspective, we treat the network as a sensor measuring the true class distribution $\mathbf{y}_i$ of the external input $\mathbf{x}_i$, an interpretation similar to established frameworks in that it views neural network training as Recursive Bayesian Estimation \citep{singhal1988training}. Although the sensor evolves during training, making the measurement noise non-stationary, this dynamic is consistent with adaptive filtering and probabilistic self-training frameworks (e.g., Expectation-Maximization). Importantly, model predictions $\hat{\mathbf{y}}_i^t$ can be viewed as independent evidence, since they constitute measurements as read by the sensor $f$, conditioned on external input $\mathbf{x}_i$. To formalize this, we introduce a latent class variable $z_i \in \{1,\ldots,k\}$ for each sample $\mathbf{x}_i$ and model its distribution as categorical, 
\begin{equation}
    \mathrm{z}_i \sim \text{Cat}(\mathbf{y}_i).
\end{equation}
where $\mathbf{y}_i = [y_{i,1}, y_{i,2}, \dots, y_{i,k}]$ represents the probabilities for each of the $k$ classes.

To express prior beliefs about $\mathbf{y}_i$, we use a Dirichlet distribution $\mathbf{y}_i \sim \text{Dir}(\bm\alpha_i)$ for its conjugacy and its construction from normalized independent Gamma variables (if $v_{i,j} \sim \text{Gamma}(\alpha_{i,j}, \lambda)$ then Dirichlet variates are obtained by the closure $y_{i,j} = v_{i,j} / \sum_k v_{i,k})$. This formulation allows us to interpret the accumulated evidence for the classes as independent, so that when the input $\mathbf{x}_i$ triggers a non-zero prediction for a secondary class (e.g., ``cat'' features in a ``dog'' image), the component for that class receives an independent sensor update. Assuming that the model learns to extract relevant features, the predictions capture semantic knowledge rather than hallucinations. We encode prior belief in the labeled class by letting
\begin{equation}
    \alpha^0_{i,j} = 
        \begin{cases}
        c, & \text{if } j = \underset{l}{\mathrm{argmax}} \ \mathrm{y}^0_{i,l}, \\
        \epsilon , & \text{otherwise,}
        \end{cases}
\end{equation}
where $\epsilon \ll c$. For sufficiently small $\epsilon$, the prior implies $p(\mathrm{z}_{i}|\bm\alpha^0_i)_j \approx 1$ for the labeled class.

At each epoch $t$, the model outputs a prediction \(\hat{\mathbf{y}}_i^t\) which we treat as a noisy measurement to update our beliefs. We assume the likelihood
\begin{equation}
    p\left(\hat{\mathbf{y}}_i |\mathbf{y}_i\right) \propto \prod_{j=1}^k y_{i,j}^{\hat{y}_{i,j}},
\end{equation} 
which generalizes the categorical likelihood to fractional evidence, and is conjugate to the prior \citep{bishop2006pattern}. This likelihood treats the prediction as partial evidence, where higher predicted probabilities correspond to stronger observations.

Formally, the posterior distribution is
\begin{equation}
    \mathbf{y}_i | \hat{\mathbf{y}}_i^{t}, \bm\alpha_i^{t-1}  \sim \text{Dir}\left(\bm\alpha_i^t \right) , \quad \text{where} \quad \bm\alpha_i^t=\bm\alpha^{t-1}_i +\hat{\mathbf{y}}_i^{t},
\end{equation}
accumulating noisy measurements into sample-specific distributions over class probabilities $\mathbf{y}$. To predict the distribution over the labels, we use the posterior predictive distribution. However, because the quality of measurements is expected to improve during training (as the model learns), the measurement noise is non-stationary. Standard Bayesian updating would weight early, noisier observations equally to later, more accurate ones. To address this, we adopt a discounted Bayesian model \citep{west2006chapter7}, effectively forgetting old evidence to adapt to the improving network. For a discounting factor $\gamma \in [0,1]$, we have $\bm\alpha_i^t=\gamma\bm\alpha^{t-1}_i +\hat{\mathbf{y}}_i^{t}$, yielding the update rules
\begin{align}
\mathbf{y}^{t}_{i} = \frac{\gamma A_i^{t-1}}{\gamma A_i^{t-1}+1}\mathbf{y}_i^{t-1} + \left(1 - \frac{\gamma A_i^{t-1}}{\gamma  A_i^{t-1}+1}\right)\hat{\mathbf{y}}_i^{t}, \quad A_i^{t} = \gamma A_i^{t-1} + 1, \quad 
\label{eq:bayesian_update}
\end{align}
for $A_i^{t}=\sum_{j=1}^k \alpha_{ij}^t$. 

From the perspective of Bayesian filtering, discounting the concentration parameters $\bm\alpha_i^{t-1}$ constitutes the prediction step, increasing the variance of the distribution, while adding the new evidence $\hat{\mathbf{y}}_i^{t}$ forms the update step \cite{sarkka2023bayesian}.

Intuitively, the Dirichlet parameters accumulate the model’s belief about how much each class is supported by the input $\mathbf{x}_i$. Discounting ensures that more recent predictions are weighted more heavily than earlier, likely worse, ones. The prior dominates early in training when variance is high, but predictions provide stronger evidence as training progresses.

\subsection{Relationship between DPS and other methods}

Related methods can be interpreted as special cases of DPS. Conventional training corresponds to taking 
$c \to \infty$, which fixes the target distribution and prevents Bayesian updating. In this limit, the framework reduces to standard distribution matching, as KL-divergence and cross-entropy are equivalent under one-hot targets.

If initialized at its fixed point $A_i^0=\frac{1}{1-\gamma}$, the recurrence $A_i^t=A_i^0=\frac{1}{1-\gamma}$ gives the 
EMA update 
\begin{equation}
\mathbf{y}^{t}_{i} = \gamma \mathbf{y}_i^{t-1} + (1 - \gamma) \hat{\mathbf{y}}_i^{t}.
\end{equation}
\begin{algorithm}[ht!]
\caption{Deep Probabilistic Supervision (DPS)}
\label{algo:refine_labels}
\textbf{Input:} Training set $\mathcal{D} = \{(\mathbf{x}_i, \mathbf{y}^0_i)\}_{i=1}^{n}$\\
\textbf{Model:} Neural network $f$ with parameters $\bm\theta^0$, optimizer $h$, loss function $\ell$\\
\textbf{Parameters:} Number of epochs $T$, 
Dirichlet prior  $\bm\alpha^0_i = [\alpha^0_{i,1}, \dots, \alpha^0_{i,k}]$, discount factor $\gamma$.

\textbf{Initialize:}
$A_i^{0}=\sum_{j=1}^k \alpha_{ij}^0$

\begin{algorithmic}
\For{epoch $t \gets 1$ to $T$}
    \For{mini-batch $B \subseteq \mathcal{D}$}
            \State $\hat{\mathbf{y}}_i^{t} \gets f(\mathbf{x}_i; \bm\theta), \quad \forall i \in B$ 
        \State $\mathcal{L} \gets \frac{1}{|B|} \sum_{i \in B} \ell\big(\hat{\mathbf{y}}_i^{t}, \mathbf{y}_i^{t-1}\big)$ 
        \State $\bm{\theta}^t \gets  h\left(\bm\theta^{t-1},\nabla_{\bm\theta^{t-1}} \mathcal{L}\right)$ 
        \State $\mathbf{y}^{t}_{i} \gets \frac{\gamma A_i^{t-1}}{\gamma A_i^{t-1}+1}\mathbf{y}_i^{t-1} + \left(1 - \frac{\gamma A_i^{t-1}}{\gamma  A_i^{t-1}+1}\right)\hat{\mathbf{y}}_i^{t}$ 
        \State $A_i^{t} \gets \gamma A_i^{t-1} + 1, \quad \forall i \in B$ 
    \EndFor
\EndFor
\State \textbf{Return} $\bm\theta^t$ 
\end{algorithmic}
\end{algorithm}

Furthermore,  
\begin{equation}
    A_i^t = \gamma A_i^{t-1}+1 = \gamma^t A_i^{0} +  \sum_{j=0}^{t-1}\gamma^j=A_i^{0}\gamma^t + \frac{1-\gamma^t}{1-\gamma} 
\end{equation}
so in the limit we have that 
\begin{equation}
    \lim_{t\to\infty} A_i^t = \frac{1}{1-\gamma},
\end{equation}
i.e. the weight of new observations converges to $1-\gamma$ exponentially. In the case of non-zero $\epsilon$, and if we initialize at the fixed point $A_i^0=\frac{1}{1-\gamma}$ for $\gamma$ close to $1$, DPS approximates label smoothing \citep{szegedy2016rethinking}.

Similarly, PS-KD \citep{kim2021self} is recovered by setting $\gamma = 0$, so that only the most recent prediction contributes to the target. 
DLB \citep{shen2022self} performs the same Bayesian update as PS-KD but applies it at the mini-batch level rather than epoch-wise. 
Meanwhile, TE \citep{laine2016temporal} corresponds to assuming an improper prior (a zero vector) and applying an explicit weighting schedule to the accumulated evidence. Although these methods differ from DPS in that they include a supervised loss term based on one-hot labels, apply temperature scaling or modify the loss function, all of them implicitly construct a target distribution from an accumulation of past predictions. DPS makes this shared structure explicit by interpreting it as Bayesian evidence aggregation under a Dirichlet prior.

\section{Experiments}
\label{sec:experiments}
\textbf{Experimental Setup.} We evaluate ResNet \citep{he2016deep}, DenseNet \citep{huang2017densely}, and ViT \citep{dosovitskiy2020image} models on CIFAR-10 \citep{krizhevsky2009learning}, CIFAR-100 \citep{krizhevsky2009learning}, Tiny ImageNet \citep{tinyimagenet} and ImageNet \cite{russakovsky2015imagenet}. We report the average of three runs. For DPS, we set $\epsilon=0$, $\gamma = 0.95$, and use $c=1000$ and $c=50$ for the CNNs and ViTs, respectively, on all datasets except for ImageNet. On ImageNet, we set $\epsilon=0.05$, $\gamma = 0.99$, and use $c=1000$. The experimental setup is described in more detail in the appendix, including the hyperparameters used for the baselines.

\subsection{Classification Results}
\textbf{Methods compared.}
We benchmark DPS against conventional training (baseline), Temporal Ensembling (TE) \citep{laine2016temporal}, self-Distillation
from Last mini-Batch (DLB) \citep{shen2022self} and Progressive Self-Knowledge Distillation (PS-KD) \citep{kim2021self}. We do not compare with methods such as \citep{zhang2019your, zhang2022self}, as our focus is on modifying label distributions rather than model architectures. An ensemble of three models is included for reference, whose knowledge is also distilled into a single model of the same architecture, for a comparison with conventional knowledge distillation. 

\subsubsection{Generalization}
The main results are reported in Table \ref{tab:accuracy_comparison}. In all experiments, DPS improves test accuracy relative to both the baseline and related methods, surpassing conventional knowledge distillation and approaching ensemble performance.

The performance gains are most pronounced on CIFAR-100 and TinyImageNet. For ResNet and DenseNet, DPS boosts accuracy by about 3 percentage points (pp) over the baseline and by more than 1 pp over the best-performing related methods on both datasets. On CIFAR-10, where baseline performance is high, the improvements are more modest. Similarly, for ViT-B/16, we observe small but consistent gains across all datasets. On ImageNet, DPS improves accuracy for ResNet-152 by about 1.5 and 0.7 pp, and for DenseNet-264 by about 2.2 and 1.0 pp, over the baseline and best-performing related method, respectively. 

\begin{table*}[ht!]
\centering
\caption{\textbf{Test set accuracy on CIFAR-10, CIFAR-100, TinyImageNet and ImageNet.} The models are trained using standard training (baseline), related methods \citep{laine2016temporal,shen2022self,kim2021self}, and the proposed method (DPS). The best results are highlighted in bold.}
\footnotesize 
\setlength{\tabcolsep}{2.0pt} 
\begin{tabular}{@{}llccccc|cc@{}}
\toprule
Dataset     & Model        & Baseline (\%)                        & TE  (\%)                        & DLB  (\%)                       & PS-KD (\%)                       & DPS (\%)                           & KD  (\%)                        & Ens. (\%) \\ \midrule
\multirow{3}{*}{CIFAR-10} 
            & ResNet-18    & $94.39{\scriptstyle\pm0.18}$         & $94.52{\scriptstyle\pm0.08}$    & $94.52{\scriptstyle\pm0.11}$    & $94.56{\scriptstyle\pm0.06}$     & $\mathbf{94.88}\,{\scriptscriptstyle{\pm0.06}}$ & $94.73{\scriptstyle\pm0.18}$    & $95.33$   \\
            & DenseNet-121 & $94.86{\scriptstyle\pm0.13}$         & $95.20{\scriptstyle\pm0.13}$    & $95.15{\scriptstyle\pm0.15}$    & $95.10{\scriptstyle\pm0.04}$     & $\mathbf{95.38}\,{\scriptscriptstyle{\pm0.13}}$ & $95.28{\scriptstyle\pm0.09}$    & $95.67$   \\
            & ViT-B/16     & $98.28{\scriptstyle\pm0.02}$         & $98.34{\scriptstyle\pm0.05}$    & $98.39{\scriptstyle\pm0.03}$    & $98.37{\scriptstyle\pm0.08}$     & $\mathbf{98.44}\,{\scriptscriptstyle{\pm0.01}}$ & $98.32{\scriptstyle\pm0.02}$    & $98.56$   \\ \midrule
\multirow{3}{*}{CIFAR-100} 
            & ResNet-50    & $75.82{\scriptstyle\pm0.41}$         & $76.07{\scriptstyle\pm0.40}$    & $77.22{\scriptstyle\pm0.23}$    & $77.71{\scriptstyle\pm0.20}$     & $\mathbf{79.09}\,{\scriptscriptstyle{\pm0.11}}$ & $77.69{\scriptstyle\pm0.08}$    & $79.29$   \\
            & DenseNet-169 & $76.63{\scriptstyle\pm0.28}$         & $76.33{\scriptstyle\pm0.11}$    & $78.43{\scriptstyle\pm0.09}$    & $77.88{\scriptstyle\pm0.04}$     & $\mathbf{79.47}\,{\scriptscriptstyle{\pm0.12}}$ & $78.72{\scriptstyle\pm0.18}$    & $79.83$   \\
            & ViT-B/16     & $89.16{\scriptstyle\pm0.05}$         & $89.11{\scriptstyle\pm0.23}$    & $88.72{\scriptstyle\pm0.04}$    & $89.36{\scriptstyle\pm0.11}$     & $\mathbf{89.54}\,{\scriptscriptstyle{\pm0.13}}$ & $89.35{\scriptstyle\pm0.11}$    & $90.27$   \\ \midrule
\multirow{3}{*}{TinyImageNet} 
            & ResNet-101   & $64.22{\scriptstyle\pm0.21}$         & $64.45{\scriptstyle\pm0.12}$    & $65.83{\scriptstyle\pm0.28}$    & $65.65{\scriptstyle\pm0.12}$     & $\mathbf{67.41}\,{\scriptscriptstyle{\pm0.31}}$ & $66.36{\scriptstyle\pm0.16}$    & $69.78$   \\
            & DenseNet-201 & $64.53{\scriptstyle\pm0.38}$         & $64.60{\scriptstyle\pm0.19}$    & $66.66{\scriptstyle\pm0.20}$    & $66.11{\scriptstyle\pm0.08}$     & $\mathbf{67.74}\,{\scriptscriptstyle{\pm0.04}}$ & $66.94{\scriptstyle\pm0.09}$    & $69.44$   \\
            & ViT-B/16     & $88.99{\scriptstyle\pm0.20}$         & $89.02{\scriptstyle\pm0.23}$    & $89.29{\scriptstyle\pm0.13}$    & $89.32{\scriptstyle\pm0.13}$     & $\mathbf{89.65}\,{\scriptscriptstyle{\pm0.17}}$ & $89.16{\scriptstyle\pm0.10}$    & $90.23$   \\ \midrule
\multirow{2}{*}{ImageNet}   
            & ResNet-152   & $78.40{\scriptstyle\pm0.16}$         & $78.46{\scriptstyle\pm0.10}$    & $78.54{\scriptstyle\pm0.14}$    & $79.18{\scriptstyle\pm0.06}$     & $\mathbf{79.88}\,{\scriptscriptstyle{\pm0.11}}$ & $79.73{\scriptstyle\pm0.05}$    & $80.35$   \\ 
            & DenseNet-264   & $76.55{\scriptstyle\pm0.02}$         & $76.57{\scriptstyle\pm0.07}$    & $77.20{\scriptstyle\pm0.07}$    & $77.83{\scriptstyle\pm0.02}$     & $\mathbf{78.79}\,{\scriptscriptstyle{\pm0.12}}$ & $78.56{\scriptstyle\pm0.06}$    & $79.69$   \\\bottomrule
\end{tabular}%
\label{tab:accuracy_comparison}
\end{table*}

\subsubsection{Calibration}
\begin{wrapfigure}{r}{0.53\textwidth}
\vspace{-20pt}
  \centering
    \begin{subfigure}[b]{0.326\linewidth}
    \centering
    \includegraphics[width=\linewidth]{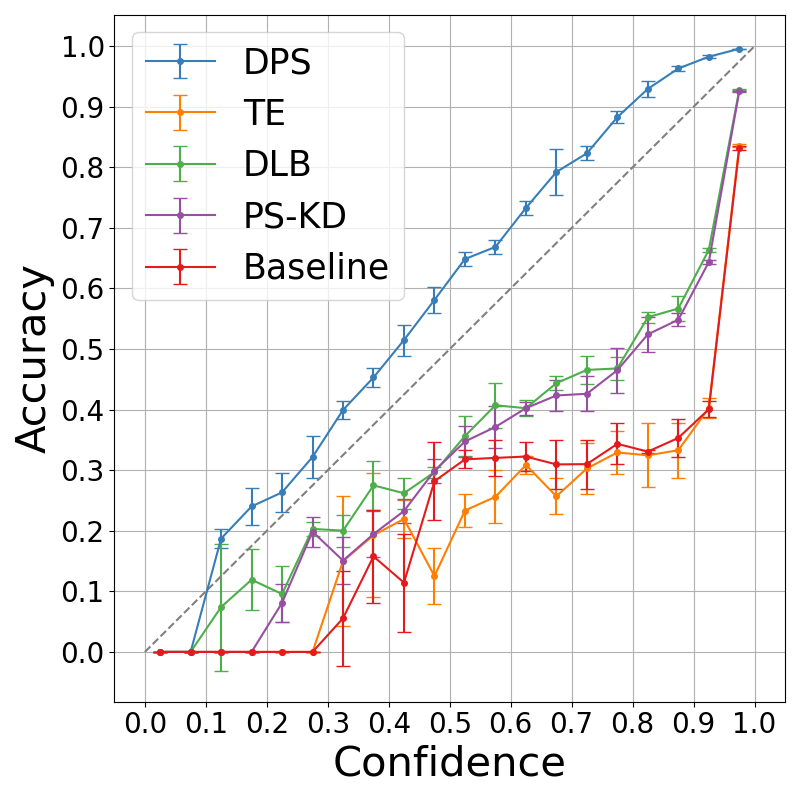}
    \caption{ResNet-50}
  \end{subfigure}
  \hfill
  \begin{subfigure}[b]{0.326\linewidth}
    \centering
    \includegraphics[width=\linewidth]{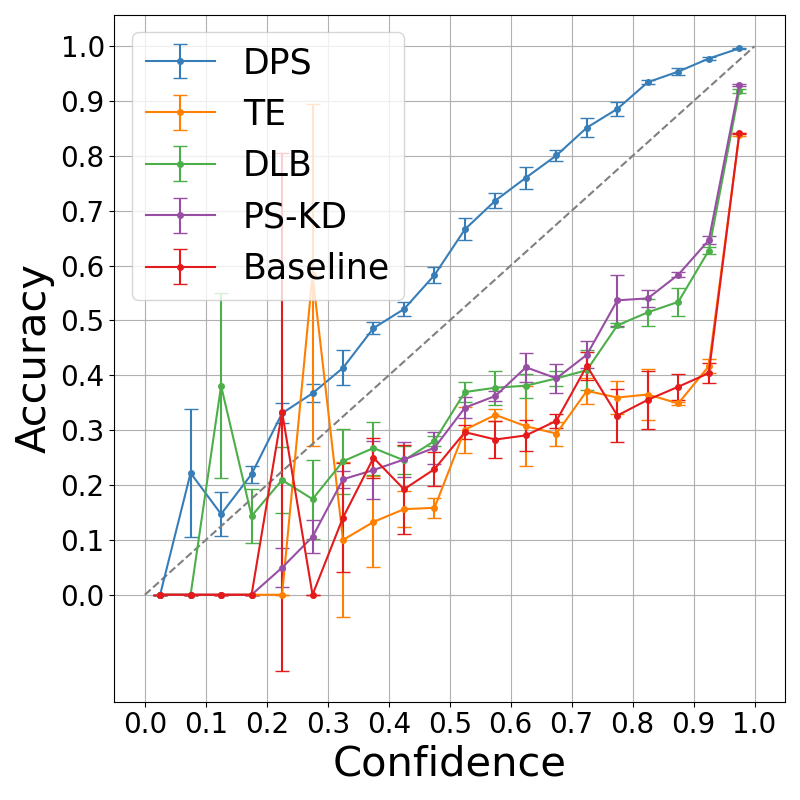}
    \caption{DenseNet-169}
  \end{subfigure}
  \hfill
  \begin{subfigure}[b]{0.326\linewidth}
    \centering
    \includegraphics[width=\linewidth]{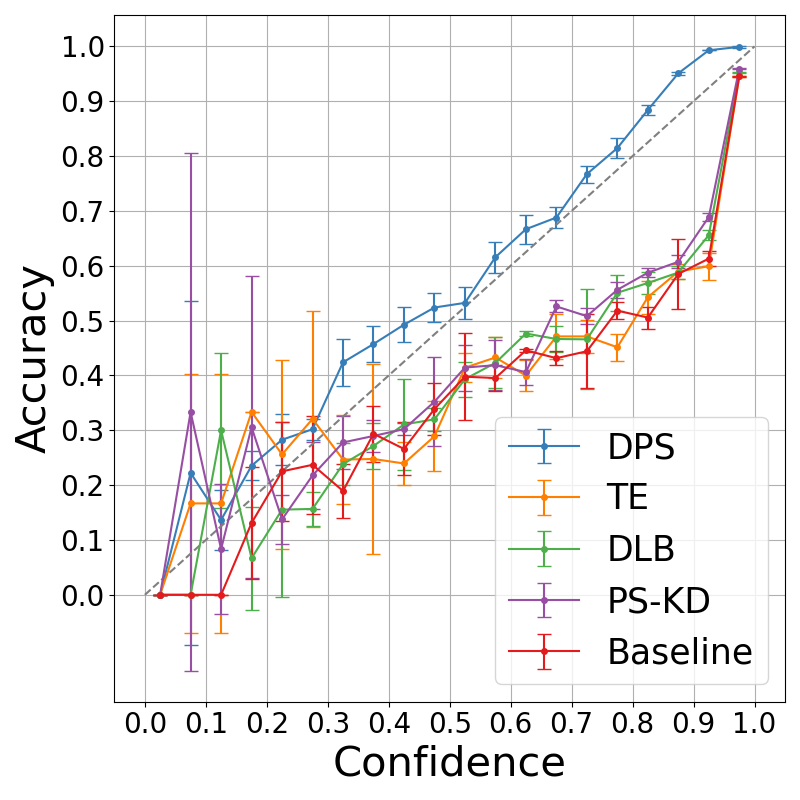}
    \caption{ViT-B}
  \end{subfigure}

  \caption{\textbf{Reliability Diagrams for CIFAR-100 Models.} A curve closer to the diagonal indicates better calibration. The models are trained using standard training (baseline), related methods \citep{laine2016temporal,shen2022self,kim2021self}, and the proposed method (DPS).} 
  \label{fig:calibration_plots}
  \vspace{-40pt}
\end{wrapfigure}

Given the importance of calibrated probability estimates in many tasks, we evaluate the calibration of models trained on CIFAR-100 using Expected Calibration Error (ECE) \citep{naeini2015obtaining} and Negative Log Likelihood (NLL). The results are included in Table \ref{tab:cifar100_calibration_methods}, where DPS demonstrates superior calibration compared to related methods. Figure~\ref{fig:calibration_plots} includes reliability diagrams where DPS’s curves lie closest to the diagonal line, indicating better calibration.
Furthermore, we compare our framework with more calibration methods in Table \ref{tab:calibration_wrn} (Appendix), where DPS achieves state-of-the-art calibration across all metrics.

\begin{wrapfigure}{r}{0.53\textwidth}
\centering
\captionof{table}{\textbf{ECE and NLL on CIFAR-100.} The models are trained using standard training (baseline), related methods \citep{laine2016temporal,shen2022self,kim2021self}, and the proposed method (DPS). The best results are highlighted in bold.}
\setlength{\tabcolsep}{2.0pt} 
    \resizebox{\linewidth}{!}{
    \begin{tabular}{@{}llccccc@{}}
\toprule
Architecture & Metric & Baseline & TE & DLB & PS-KD & DPS \\ \midrule
\multirow{2}{*}{ResNet-50} 
& ECE (\%) & $20.41{\scriptstyle\pm0.45}$ & $20.21{\scriptstyle\pm0.34}$ & $11.82{\scriptstyle\pm0.28}$ & $12.41{\scriptstyle\pm0.17}$ & $\mathbf{7.17}\,{\scriptscriptstyle{\pm0.40}}$ \\
& NLL      & $2.94{\scriptstyle\pm0.04}$  & $2.91{\scriptstyle\pm0.04}$  & $1.09{\scriptstyle\pm0.01}$  & $1.09{\scriptstyle\pm0.02}$  & $\mathbf{0.77}\,{\scriptscriptstyle{\pm0.00}}$  \\ \midrule
\multirow{2}{*}{DenseNet-169}
& ECE (\%) & $19.33{\scriptstyle\pm0.24}$ & $19.63{\scriptstyle\pm0.16}$ & $12.42{\scriptstyle\pm0.09}$ & $12.35{\scriptstyle\pm0.10}$ & $\mathbf{7.67}\,{\scriptscriptstyle{\pm0.10}}$ \\
& NLL      & $2.59{\scriptstyle\pm0.03}$  & $2.63{\scriptstyle\pm0.01}$  & $1.06{\scriptstyle\pm0.01}$  & $1.08{\scriptstyle\pm0.01}$  & $\mathbf{0.75}\,{\scriptscriptstyle{\pm0.00}}$ \\ \midrule
\multirow{2}{*}{ViT-B/16}
& ECE (\%) & $7.53{\scriptstyle\pm0.15}$  & $7.38{\scriptstyle\pm0.26}$  & $6.89{\scriptstyle\pm0.08}$  & $5.93{\scriptstyle\pm0.26}$  & $\mathbf{5.89}\,{\scriptscriptstyle{\pm0.09}}$ \\
& NLL      & $0.56{\scriptstyle\pm0.02}$  & $0.55{\scriptstyle\pm0.01}$  & $0.52{\scriptstyle\pm0.00}$  & $0.44{\scriptstyle\pm0.01}$  & $\mathbf{0.42}\,{\scriptscriptstyle{\pm0.00}}$ \\ \bottomrule
\end{tabular}
}

\label{tab:cifar100_calibration_methods}
\vspace{-15pt}

\end{wrapfigure}

The improvements are significant compared to the baseline, with DPS reducing ECE by more than 60\% and NLL by over 70\% for the convolutional networks. Relative to the best-performing related method, DPS lowers ECE by nearly 40\% and NLL by about 29\%. For ViT-B/16, the gains are more modest, with a 22\% reduction in ECE and a 25\% reduction in NLL over the baseline, alongside small but consistent improvements over related methods. As discussed in Section \ref{sec:ablation}, careful hyperparameter selection can further improve DPS’s calibration, e.g. reaching an ECE of 1.33\% for ResNet-50 on CIFAR-100.

\subsubsection{Robustness}
To assess our framework's performance under more realistic conditions, we evaluate the models trained on CIFAR-10 on corrupted and perturbed images, and introduce symmetric and asymmetric label noise.

\textbf{Corruptions and perturbations.} We evaluate the robustness to corruptions and perturbations of DPS and related methods by evaluating the models trained on CIFAR-10 on CIFAR-10-C and CIFAR-10-P \citep{hendrycks2019benchmarking}. The results are included in Table \ref{tab:cifar10cp_results}, where we report test set accuracy and mean Flip Probability (mFP) for CIFAR-10-C and CIFAR-10-P, respectively. DPS yields the highest accuracy and the lowest mFP for all models.

On CIFAR-10-C, DPS improves robustness across all architectures, raising accuracy by about $1.5$ pp for both ResNet-18 and DenseNet-121 and by $0.4$ pp for ViT-B over the baseline, slightly surpassing the best related methods. On CIFAR-10-P, DPS achieves the lowest mean flip probability for all models, reducing mFP by around $10\%$ compared to the best-performing related methods.

\begin{wrapfigure}{r}{0.30\textwidth}
\vspace{-20pt}
    \centering
    \includegraphics[width=\linewidth]{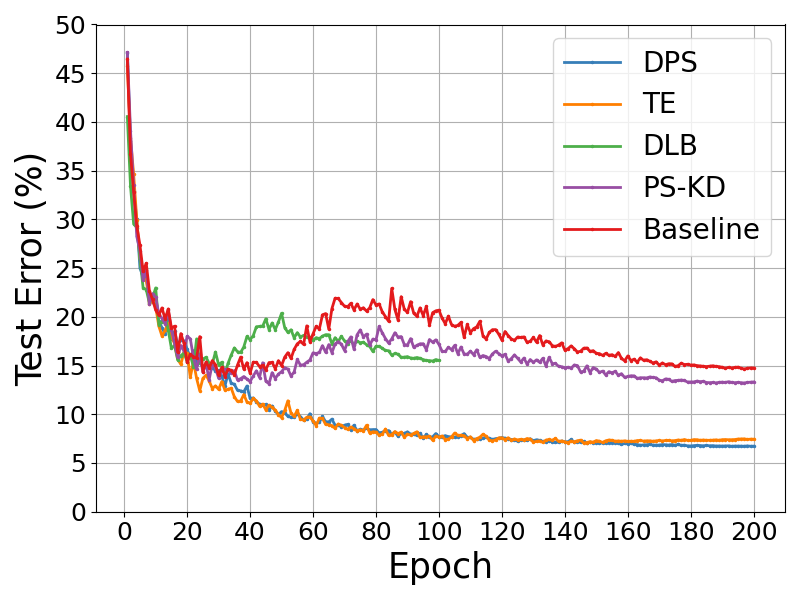}  
    \caption{\textbf{Test Error Over Epochs Under 20\% Label Noise for ResNet-18 on CIFAR-10.} The models are trained using standard training (baseline), related methods \citep{laine2016temporal,shen2022self,kim2021self}, and the proposed method (DPS).}
    \label{fig:dd} 
\vspace{-20pt}
\end{wrapfigure}

\textbf{Label noise.} We generate symmetric label noise by randomly reassigning a percentage of all labels, and asymmetric noise using the class-dependent definition from prior work \citep{tanaka2018joint}. Figure~\ref{fig:dd} shows test error over epochs for the previously mentioned self-distillation methods under 20\% label noise for ResNet-18 trained on CIFAR-10. DPS and TE obtain the highest accuracy under label noise, while the methods that rely more on the hard targets perform worse. Notably, DPS flattens the characteristic double descent curve observed for other methods, indicating a more robust learning process.

\begin{table*}[ht!]
\centering
\caption{\textbf{Performance on CIFAR-10-C and CIFAR-10-P.} Accuracy and Mean Flip Probability of DPS, baseline, and related methods \citep{laine2016temporal,shen2022self,kim2021self} under corruptions and perturbations. The best results for each metric are highlighted in bold.}
\label{tab:cifar10cp_results}
\footnotesize
\setlength{\tabcolsep}{2.0pt} 
\resizebox{\linewidth}{!}{\begin{tabular}{lcccccccccc}
\toprule
\multirow{2}{*}{Model} & \multicolumn{5}{c}{CIFAR-10-C (Acc., \%)} & \multicolumn{5}{c}{CIFAR-10-P (mFP, \%)} \\
\cmidrule(lr){2-6} \cmidrule(lr){7-11}
& Baseline & TE & DLB & PS-KD & DPS & Baseline & TE & DLB & PS-KD & DPS \\
\midrule
ResNet-18    & $73.31{\scriptstyle\pm0.31}$ & $73.22{\scriptstyle\pm0.27}$ & $73.94{\scriptstyle\pm0.31}$ & $74.28{\scriptstyle\pm0.50}$ & $\textbf{74.88}\,{\scriptscriptstyle\pm0.28}$ & $7.20{\scriptstyle\pm0.21}$ & $7.23{\scriptstyle\pm0.04}$ & $7.08{\scriptstyle\pm0.06}$ & $7.10{\scriptstyle\pm0.28}$ & $\textbf{6.09}\,{\scriptscriptstyle\pm0.08}$ \\
DenseNet-121 & $73.36{\scriptstyle\pm0.32}$ & $73.09{\scriptstyle\pm0.86}$ & $74.81{\scriptstyle\pm0.23}$ & $74.30{\scriptstyle\pm0.07}$ & $\textbf{74.85}\,{\scriptscriptstyle\pm0.32}$ & $7.49{\scriptstyle\pm0.20}$ & $7.46{\scriptstyle\pm0.18}$ & $7.06{\scriptstyle\pm0.22}$ & $7.21{\scriptstyle\pm0.16}$ & $\textbf{6.39}\,{\scriptscriptstyle\pm0.11}$ \\
ViT-B        & $91.17{\scriptstyle\pm0.07}$ & $91.31{\scriptstyle\pm0.22}$ & $91.20{\scriptstyle\pm0.26}$ & $91.38{\scriptstyle\pm0.24}$ & $\textbf{91.57}\,{\scriptscriptstyle\pm0.12}$ & $2.53{\scriptstyle\pm0.04}$ & $2.40{\scriptstyle\pm0.03}$ & $2.45{\scriptstyle\pm0.05}$ & $2.35{\scriptstyle\pm0.06}$ & $\textbf{2.11}\,{\scriptscriptstyle\pm0.07}$ \\
\bottomrule
\end{tabular}}
\vspace{-12pt}
\end{table*}

\begin{wrapfigure}{r}{0.63\textwidth}
\centering
    \caption{\textbf{Test set accuracy under symmetric and asymmetric label noise.} Performance of lightweight methods (top, ResNet-18) \citep{szegedy2016rethinking,wang2019symmetric,zhang2017mixup,laine2016temporal} and state-of-the-art methods (bottom, PreAct ResNet-18) \citep{feng2023ot,chang2023csot,xu2025revisiting} on CIFAR-10. The best results are highlighted in bold.}
    \label{tab:noise_combined}
    
    \setlength{\tabcolsep}{2.0pt} 
    \footnotesize
        \resizebox{\linewidth}{!}{
        \begin{tabular}{lcccccc}
        \toprule
        \multirow{2}{*}{Method} & \multicolumn{3}{c}{Symmetric} & \multicolumn{3}{c}{Asymmetric } \\
        \cmidrule(lr){2-4} \cmidrule(lr){5-7}
                               & 20\% & 50\% & 80\% & 10\% & 30\% & 40\% \\
        \midrule
        Baseline & $86.76{\scriptstyle\pm0.37}$ & $81.49{\scriptstyle\pm0.26}$ & $63.63{\scriptstyle\pm0.38}$ & $90.07{\scriptstyle\pm0.26}$ & $85.23{\scriptstyle\pm0.29}$ & $80.30{\scriptstyle\pm0.60}$ \\
        LS       & $87.85{\scriptstyle\pm0.10}$ & $81.49{\scriptstyle\pm0.51}$ & $64.70{\scriptstyle\pm1.17}$ & $90.35{\scriptstyle\pm0.24}$ & $86.10{\scriptstyle\pm0.59}$ & $81.96{\scriptstyle\pm0.52}$ \\
        SL       & $91.86{\scriptstyle\pm0.12}$ & $86.39{\scriptstyle\pm0.32}$ & $72.89{\scriptstyle\pm0.45}$ & $91.95{\scriptstyle\pm0.14}$ & $86.39{\scriptstyle\pm0.21}$ & $80.57{\scriptstyle\pm0.07}$ \\
        MixUp    & $89.87{\scriptstyle\pm0.18}$ & $83.38{\scriptstyle\pm0.20}$ & $69.23{\scriptstyle\pm0.56}$ & $91.78{\scriptstyle\pm0.18}$ & $88.26{\scriptstyle\pm0.58}$ & $84.49{\scriptstyle\pm0.54}$ \\
        TE       & $93.07{\scriptstyle\pm0.07}$ & $90.70{\scriptstyle\pm0.14}$ & $72.61{\scriptstyle\pm0.74}$ & $94.37{\scriptstyle\pm0.15}$ & $92.86{\scriptstyle\pm0.07}$ & $90.81{\scriptstyle\pm0.07}$ \\
        DPS      & $93.39{\scriptstyle\pm0.05}$ & $90.71{\scriptstyle\pm0.26}$ & $77.50{\scriptstyle\pm0.49}$ & $94.18{\scriptstyle\pm0.07}$ & $93.20{\scriptstyle\pm0.14}$ & $91.65{\scriptstyle\pm0.06}$ \\
        \midrule
        OT-Filter & $96.0$ & $95.3$ & $94.0$ & - & - & $95.1$ \\
        CSOT      & $96.6\,{\scriptscriptstyle\pm0.10}$ & $\mathbf{96.2}\,{\scriptscriptstyle\pm0.11}$ & $94.4\,{\scriptscriptstyle\pm0.16}$ & - & - & $95.5\,{\scriptscriptstyle\pm0.06}$ \\
        DULC      & $96.6$ & $96.0$ & $95.0$ & $96.7$ & $95.5$ & $95.2$ \\
        DPS+      & $\mathbf{96.9}\,{\scriptscriptstyle\pm0.04}$ & $\mathbf{96.2}\,{\scriptscriptstyle\pm0.06}$ & $\mathbf{95.1}\,{\scriptscriptstyle\pm0.06}$ & $\mathbf{97.0}\,{\scriptscriptstyle\pm0.12}$ & $\mathbf{96.3}\,{\scriptscriptstyle\pm0.13}$ & $\mathbf{95.6}\,{\scriptscriptstyle\pm0.25}$ \\
        \bottomrule
        \end{tabular}
        }
        \vspace{-10pt}
\end{wrapfigure}

\newpage

We compare DPS with lightweight regularization methods under noisy labels, including Label Smoothing (LS) \citep{szegedy2016rethinking}, Symmetric Cross Entropy Learning (SL) \citep{wang2019symmetric}, MixUp \citep{zhang2017mixup} and Temporal Ensembling (TE) \cite{laine2016temporal}. We report the average best obtained accuracy, which for DPS is typically close to the final accuracy, but diverges in varying amounts for the remaining methods. The results are included in upper section of Table \ref{tab:noise_combined}, where DPS yields the highest accuracy for all noise levels except under $10\%$ asymmetric noise. We hypothesize that the strong performance of DPS is due to the complete detachment from the original targets during training, which would explain the robustness under high noise levels.

To compare DPS with state-of-the-art methods for learning with noisy labels, we combine it with techniques from self-supervised learning (for more information refer to the Appendix). We restrict our analysis to single-stage, single-network approaches, as multi-stage or ensemble methods may be used to further enhance DPS's performance. We include OT-Filter \citep{feng2023ot}, Curriculum and Structure-aware Optimal Transport (CSOT) \citep{chang2023csot} and Dynamic and Uniform Label Correction (DULC) \citep{xu2025revisiting}. Like other works, we consider a PreAct ResNet-18 \citep{he2016identity}, trained with SGD for 300 epochs with a batch size of $128$, weight decay of $5e^{-4}$, and learning rate of 0.02, while reporting the average best obtained accuracy. Results are summarized in the lower section of Table \ref{tab:noise_combined}, where DPS consistently matches or outperforms existing state-of-the-art methods.

\subsubsection{Augmentation}
As data augmentation is a standard component in image classification pipelines, we examine the interaction of DPS with two widely used augmentation methods: CutOut \citep{devries2017improved} and CutMix \citep{yun2019cutmix}. CutOut masks out a patch of an image with zeros, while CutMix replaces a patch with a segment from another image and interpolates the labels. For DPS, the augmentations are applied to half of the images in each mini-batch, while the remainder is used for distillation. To compensate for the reduced frequency of label updates, we reduce $c$ by half (except for ImageNet, where we keep it at $c=1000$), while doubling the weight of new predictions ($1-\gamma$). Results are reported in Table~\ref{tab:accuracy_comparison_cutmix_cutout}, where we include DLB \citep{shen2022self} and conventional training for comparison.

\begin{table*}[ht!]
\centering
\caption{\textbf{Accuracy on CIFAR-10, CIFAR-100, and TinyImageNet with CutMix and CutOut.} The models are trained using standard training (baseline), DLB \citep{shen2022self} and the proposed method (DPS). The best results are highlighted in bold.}

\setlength{\tabcolsep}{2.0pt}
\footnotesize
\begin{tabular}{@{}ll
    ccc  
    ccc  
    @{}}
\toprule
Dataset & Model & \multicolumn{3}{c}{+ CutOut (\%)} & \multicolumn{3}{c}{+ CutMix (\%)} \\ \cmidrule(lr){3-5} \cmidrule(lr){6-8}
        &        & Baseline & DLB & DPS        & Baseline & DLB & DPS \\ \midrule
\multirow{3}{*}{CIFAR-10} 
        & ResNet-18    & $95.71{\scriptstyle\pm0.09}$ & $95.47{\scriptstyle\pm0.07}$ & $\mathbf{95.94}\,{\scriptscriptstyle\pm0.07}$  & $95.75{\scriptstyle\pm0.02}$  & $95.70{\scriptstyle\pm0.04}$ & $\mathbf{96.23}\,{\scriptscriptstyle\pm0.06}$ \\
        & DenseNet-121 & $96.09{\scriptstyle\pm0.08}$ & $95.87{\scriptstyle\pm0.12}$ & $\mathbf{96.21}\,{\scriptscriptstyle\pm0.12}$  & $96.17{\scriptstyle\pm0.20}$  & $96.34{\scriptstyle\pm0.10}$ & $\mathbf{96.84}\,{\scriptscriptstyle\pm0.08}$ \\
        & ViT-B/16     & $98.65{\scriptstyle\pm0.04}$ & $98.58{\scriptstyle\pm0.06}$ & $\mathbf{98.68}\,{\scriptscriptstyle\pm0.08}$  & $98.82{\scriptstyle\pm0.02}$  & $98.69{\scriptstyle\pm0.04}$ & $\mathbf{98.86}\,{\scriptscriptstyle\pm0.04}$ \\ \midrule
\multirow{3}{*}{CIFAR-100} 
        & ResNet-50    & $76.61{\scriptstyle\pm0.43}$ & $76.38{\scriptstyle\pm0.33}$ & $\mathbf{79.70}\,{\scriptscriptstyle\pm0.16}$  & $79.73{\scriptstyle\pm0.13}$  & $80.15{\scriptstyle\pm0.30}$ & $\mathbf{81.42}\,{\scriptscriptstyle\pm0.08}$ \\
        & DenseNet-169 & $77.10{\scriptstyle\pm0.17}$ & $77.87{\scriptstyle\pm0.16}$ & $\mathbf{80.14}\,{\scriptscriptstyle\pm0.35}$  & $79.31{\scriptstyle\pm0.27}$  & $79.93{\scriptstyle\pm0.11}$ & $\mathbf{82.18}\,{\scriptscriptstyle\pm0.14}$ \\
        & ViT-B/16     & $90.30{\scriptstyle\pm0.06}$ & $89.91{\scriptstyle\pm0.10}$ & $\mathbf{90.31}\,{\scriptscriptstyle\pm0.13}$  & $90.31{\scriptstyle\pm0.05}$  & $90.50{\scriptstyle\pm0.11}$ & $\mathbf{90.87}\,{\scriptscriptstyle\pm0.04}$ \\ \midrule
\multirow{3}{*}{TinyImageNet} 
        & ResNet-101   & $64.08{\scriptstyle\pm0.35}$ & $64.26{\scriptstyle\pm0.44}$ & $\mathbf{66.45}\,{\scriptscriptstyle\pm0.18}$  & $68.32{\scriptstyle\pm0.16}$  & $68.74{\scriptstyle\pm0.14}$ & $\mathbf{70.29}\,{\scriptscriptstyle\pm0.14}$ \\
        & DenseNet-201 & $65.58{\scriptstyle\pm0.17}$ & $64.84{\scriptstyle\pm0.34}$ & $\mathbf{67.62}\,{\scriptscriptstyle\pm0.21}$  & $66.54{\scriptstyle\pm0.60}$  & $69.22{\scriptstyle\pm0.48}$ & $\mathbf{70.63}\,{\scriptscriptstyle\pm0.30}$ \\
        & ViT-B/16     & $89.63{\scriptstyle\pm0.04}$ & $89.72{\scriptstyle\pm0.03}$ & $\mathbf{89.96}\,{\scriptscriptstyle\pm0.20}$  & $89.48{\scriptstyle\pm0.03}$  & $90.03{\scriptstyle\pm0.05}$ & $\mathbf{90.63}\,{\scriptscriptstyle\pm0.14}$ \\ \midrule
\multirow{2}{*}{ImageNet} 
        & ResNet-152   & $78.98{\scriptstyle\pm0.14}$ & $79.00{\scriptstyle\pm0.02}$ & $\mathbf{80.06}\,{\scriptscriptstyle\pm0.10}$  & $80.72{\scriptstyle\pm0.18}$  & $80.35{\scriptstyle\pm0.03}$ & $\mathbf{80.83}\,{\scriptscriptstyle\pm0.07}$ \\
& DenseNet-264   & $77.12{\scriptstyle\pm0.05}$ & $77.75{\scriptstyle\pm0.01}$ & $\mathbf{78.91}\,{\scriptscriptstyle\pm0.10}$  & $79.02{\scriptstyle\pm0.16}$  & $79.46{\scriptstyle\pm0.05}$ & $\mathbf{79.57}\,{\scriptscriptstyle\pm0.06}$\\

\bottomrule
\end{tabular}
\label{tab:accuracy_comparison_cutmix_cutout}
\end{table*}

\textbf{CutMix.} DPS consistently improves upon both the baseline and DLB across datasets and architectures. The gains are most pronounced on CIFAR-100 and TinyImageNet, where the convolutional networks improve by between $1.7\text{--}4.1$ pp over the baseline and by between $1.1\text{--}2.3$ pp over DLB. On ImageNet, DPS beats the baseline and DLB by $0.1\text{--0.6}$ pp. For ViT-B/16, DPS achieves gains of $+0.5$ and $+0.3$ pp over the baseline DLB respectively on CIFAR-100, while improving by $+1.15$ and $+0.6$ pp on TinyImageNet. 

\textbf{CutOut.} The effect of CutOut is more nuanced. On CIFAR-100, DPS consistently performs best, improving convolutional networks by more than $+3$ and $+2$ pp over the baseline and DLB, respectively. For ViT-B/16, DPS yields similar or slightly higher accuracy across datasets. While DPS surpasses the baseline on TinyImageNet, the accuracy is lower than without CutOut (cf. Table~\ref{tab:accuracy_comparison}). We speculate CutOut is over-regularizing for the complex but low resolution dataset. On ImageNet, DPS increases accuracy on ImageNet over both the baseline and DLB by $+1$ and more than $+2$ pp, for ResNet-152 and DenseNet-264, respectively.

\subsection{Analyzing Dark Knowledge}
\label{sec:emergence}
To analyze how DPS affects a model's predicted distributions, we first validate that dark knowledge emerges naturally during training, and show that DPS promotes further discovery of underlying structures in the data. To formalize the notion of dark knowledge, we decompose the output of the network, for an input $\mathbf{x}_i$ with integer label $k = \underset{l}{\mathrm{argmax}} \ \mathbf{y}_{i,l}^0$, as
\begin{equation}
    f(\mathbf{x}_i) = \bm\mu_{k} + \delta(\mathbf{x}_i).
    \label{eq:dk_def}
\end{equation}
Here, \(\bm\mu_{k}  \in \Delta^{K}\) is the expected prediction of a given class
\begin{equation}
\bm\mu_{k} = \mathbb{E}[f(X)\mid Y=k]
\label{eq:definition}
\end{equation}
and captures the inter-class component of dark knowledge. The function \(\delta: \mathbb{R}^d \to \mathbb{R}^K\) captures the sample-specific deviation from $\bm\mu_{k}$. We approximate the expectation in Equation \ref{eq:definition} using the empirical mean for a given class over the dataset, and compute $\delta(\mathbf{x}_i)=f(\mathbf{x}_i)-\bm\mu_k$. We hypothesize that the independent evidence counters resulting from the decomposition of the Dirichlet prior into independent gamma distributed random variables (Section \ref{sec:dps}) results in more nuanced predicted distributions for both classes and individual samples.

\begin{figure*}[ht!]
    \centering
    \begin{minipage}[t]{0.31\textwidth}
        \vspace{0pt} 
        \centering
        
        \begin{subfigure}{\linewidth}
            \centering
            \includegraphics[trim={0.5cm 0.7cm 0.2cm 0.4cm}, clip, width=\linewidth]{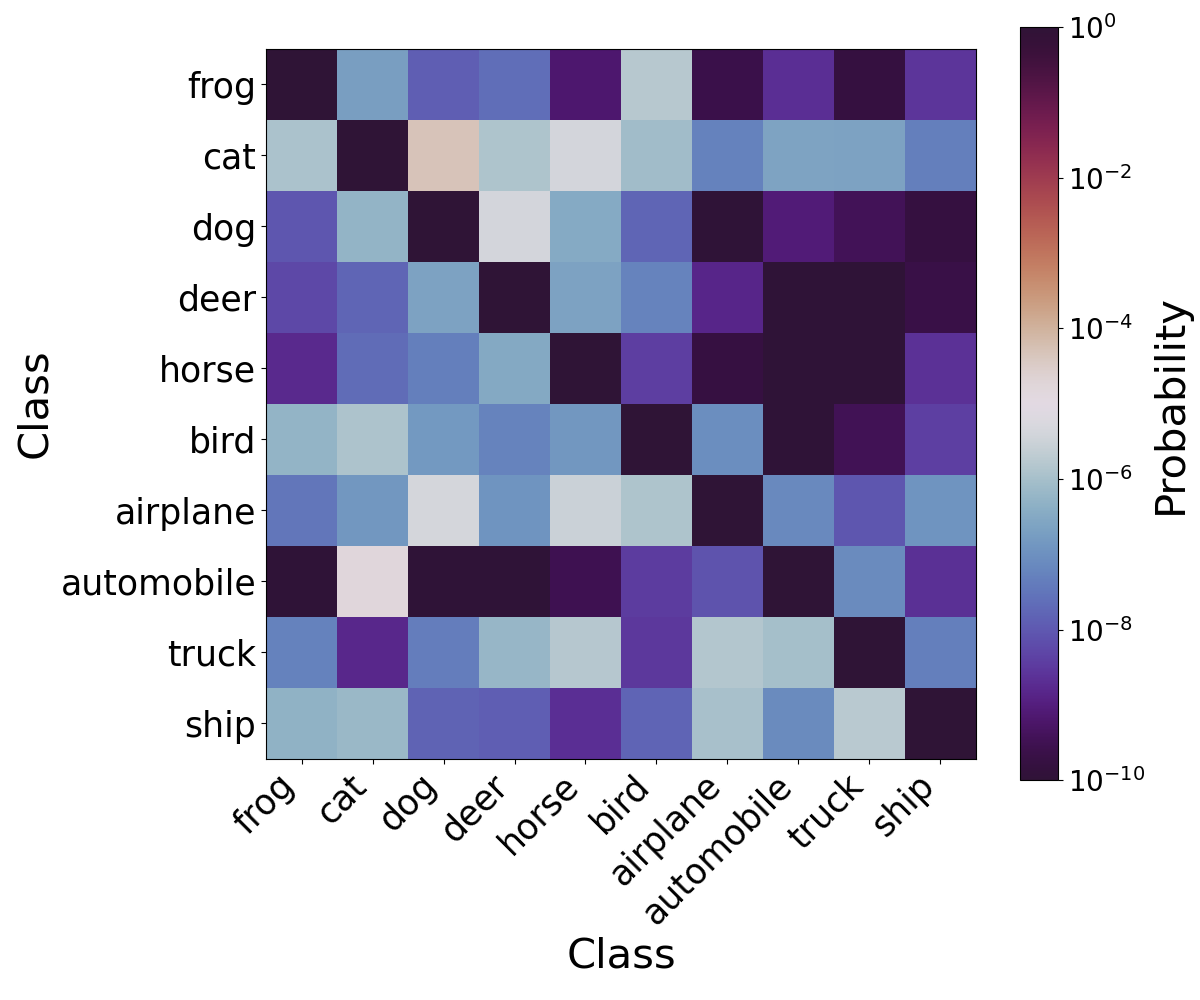}
            \caption{Conventional training}
            \label{fig:S_cifar10_conv} 
        \end{subfigure}
        
        \vspace{0.2cm}
        
        \begin{subfigure}{\linewidth}
            \centering
            \includegraphics[trim={0.5cm 0.7cm 0.2cm 0.4cm}, clip, width=\linewidth]{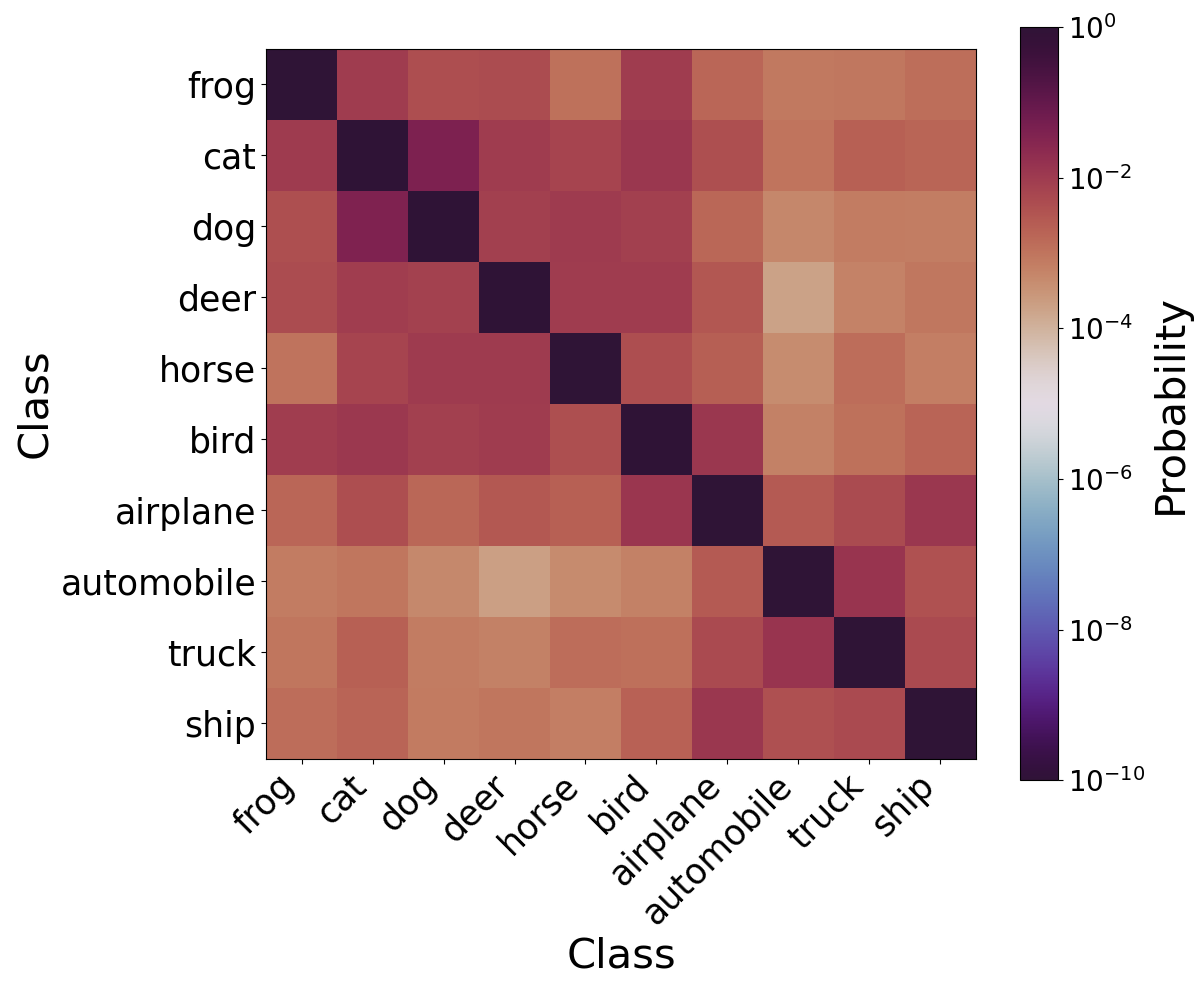}
            \caption{DPS}
            \label{fig:S_cifar10_dps}
        \end{subfigure}
    \end{minipage}
    \hfill
    \begin{minipage}[t]{0.33\textwidth}
        \vspace{0pt} 
        \centering
        
        \begin{subfigure}{\linewidth}
            \centering
            \includegraphics[width=\linewidth]{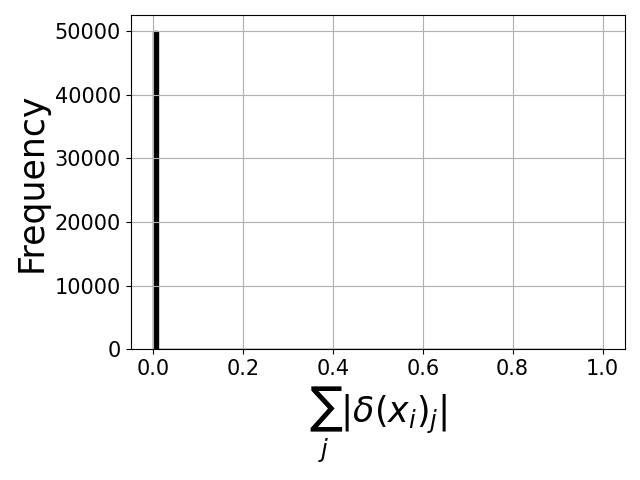}
            \caption{Conventional training}
            \label{fig:delta_cifar10_conv}
        \end{subfigure}
        
        \vspace{0.3cm}
        
        \begin{subfigure}{\linewidth}
            \centering
            \includegraphics[width=\linewidth]{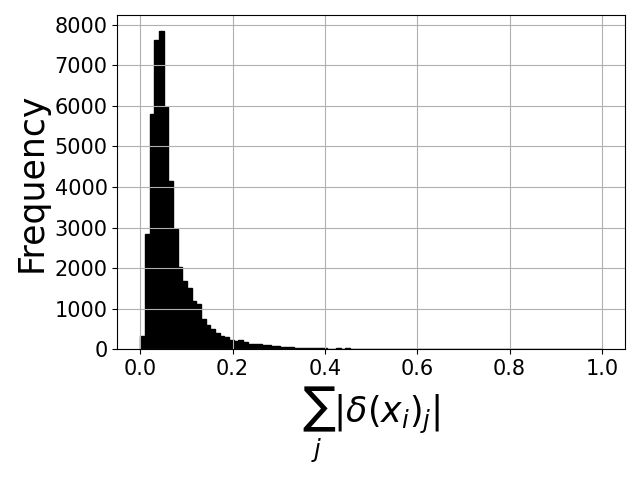}
            \caption{DPS}
            \label{fig:delta_cifar10_dps}
        \end{subfigure}
    \end{minipage}
    \hfill
    \begin{minipage}[t]{0.31\textwidth}
        \vspace{0pt} 
        \centering
        
        \begin{subfigure}{\linewidth}
            \centering
            \includegraphics[trim={0.6cm 0.1cm 1.4cm 0.3cm}, clip, width=\linewidth]{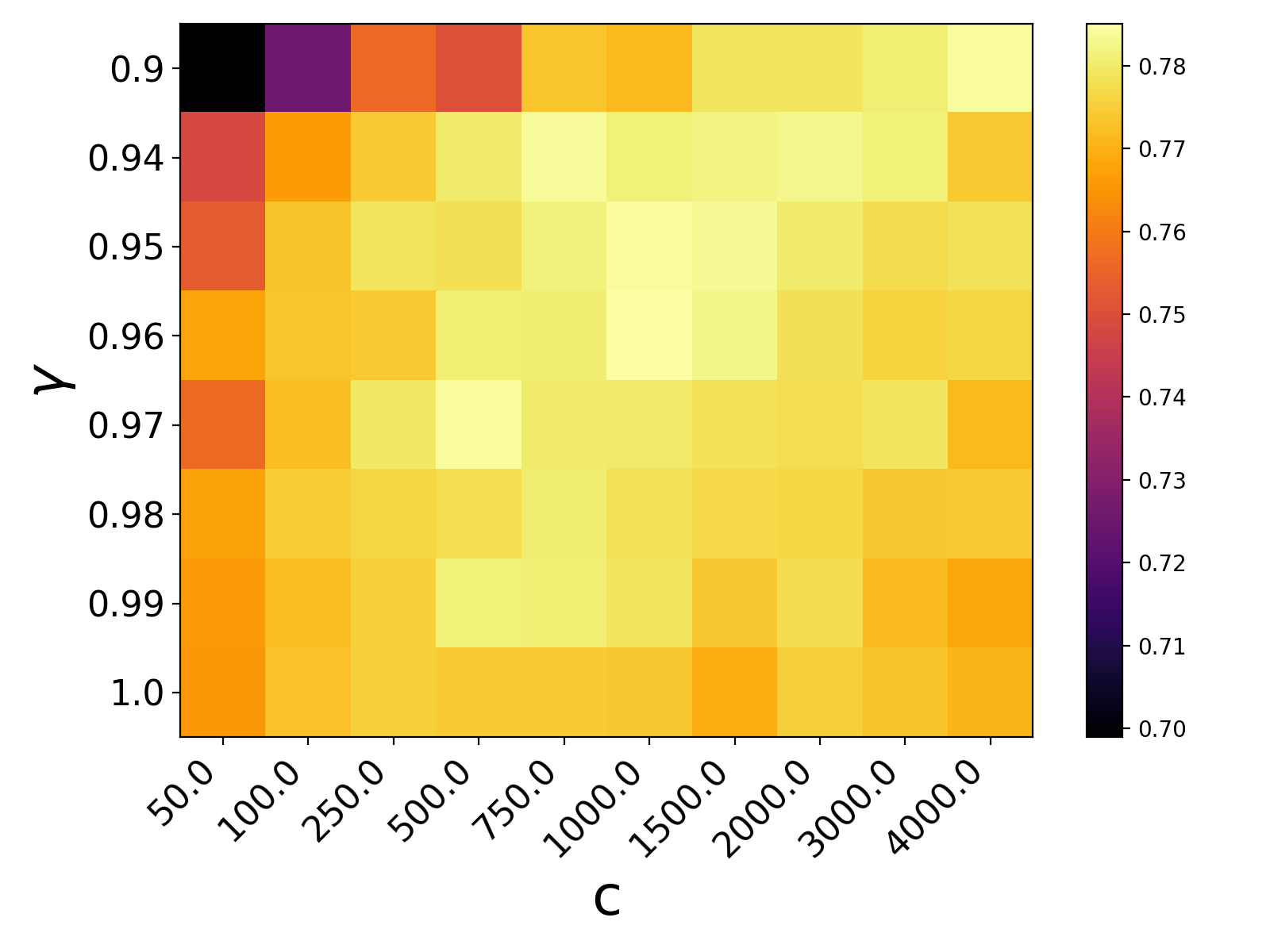}
            \caption{Accuracy}
            \label{fig:ablation_acc}
        \end{subfigure}
        
        \vspace{0.2cm}
        
        \begin{subfigure}{\linewidth}
            \centering
            \includegraphics[trim={0.6cm 0.3cm 1.4cm 0.3cm}, clip, width=\linewidth]{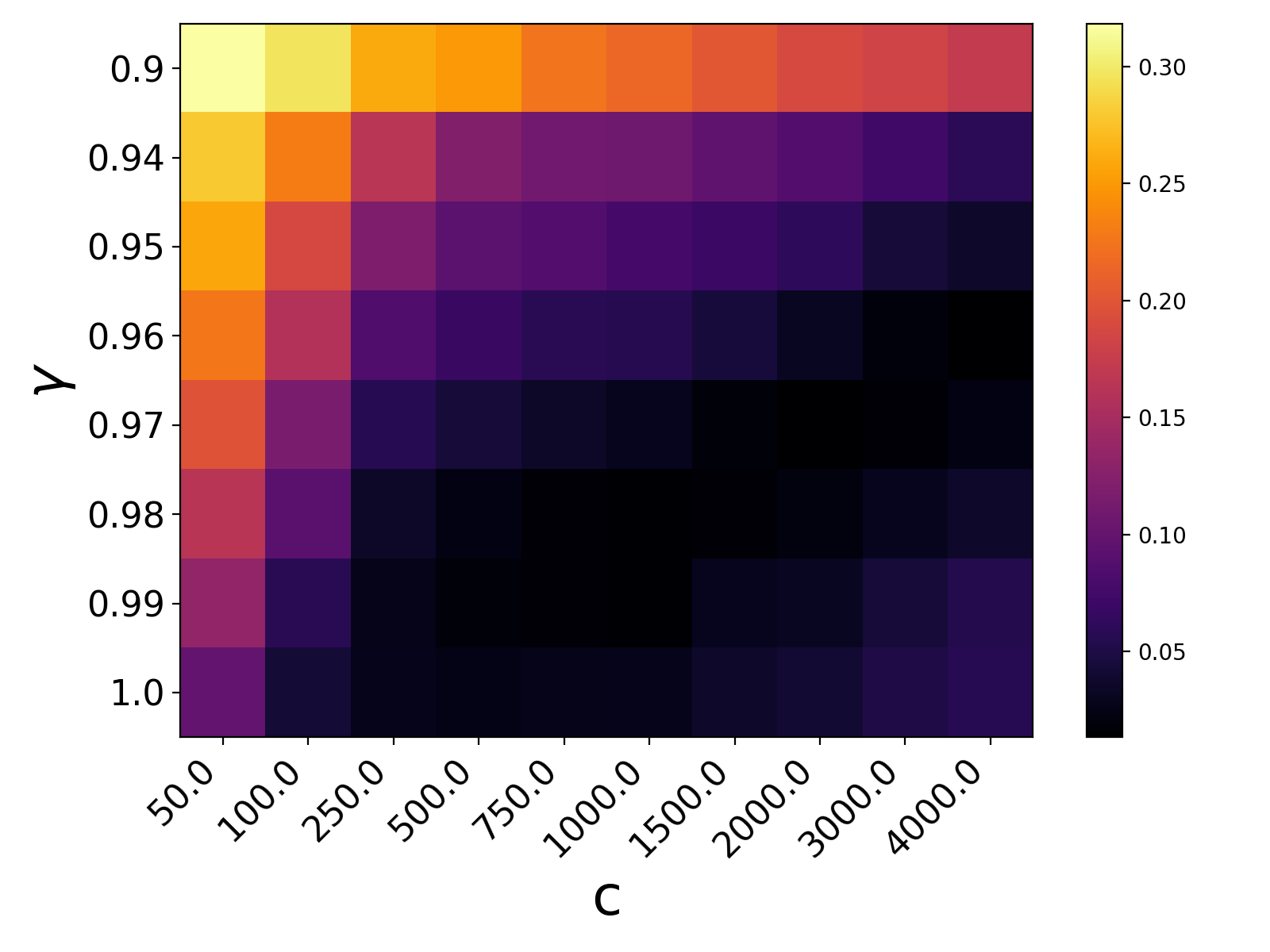}
            \caption{ECE}
            \label{fig:ablation_ece}
        \end{subfigure}
    \end{minipage}

    \caption{\textbf{Visualization of dark knowledge and ablation study.} 
    (\subref{fig:S_cifar10_conv}, \subref{fig:S_cifar10_dps}) \textbf{Inter-class component $\bm\mu$}: Semantical patterns emerge, accentuated by DPS (ResNet-18, CIFAR-10). 
    (\subref{fig:delta_cifar10_conv}, \subref{fig:delta_cifar10_dps}) \textbf{Sample-specific component}: DPS promotes learning of sample-specific information compared to conventional training. 
    (\subref{fig:ablation_acc}, \subref{fig:ablation_ece}) \textbf{Impact of $\gamma$ and $c$}: Validation accuracy and ECE sensitivity (ResNet-50, CIFAR-100).}
    
    \label{fig:full_analysis}
\end{figure*}

We visualize $\bm\mu$ and $\delta$ from a ResNet-18 trained on CIFAR-10 in Figures~\ref{fig:S_cifar10_conv}-\subref{fig:delta_cifar10_dps}. The log-scale heatmaps of $\bm\mu$ in Figures~\ref{fig:S_cifar10_conv}-\subref{fig:S_cifar10_dps} reveal that the networks learn meaningful inter-class relationships. For instance, they capture similarities between animals but also between bird and airplane, with these clusters being more prominent and the probabilities larger for DPS. The per-sample absolute deviation from $\bm\mu$ presented in Figures~\ref{fig:delta_cifar10_conv}-\subref{fig:delta_cifar10_dps} validates the emergence of $\delta$ and shows that DPS promotes learning of sample-specific information.

\subsection{Ablation}
\label{sec:ablation}    

We perform ablations on the hyperparameters $\gamma$ and $c$, analyzing their effect on accuracy and ECE for ResNet-50 on CIFAR-100. The results are plotted in Figures \ref{fig:ablation_acc} and \ref{fig:ablation_ece}. We observe high accuracy and low ECE for a range of values. Interestingly, the set of hyperparameters that minimizes ECE does not coincide exactly with the values that maximizes accuracy. 

The lowest ECE is achieved for $\gamma=0.97$ and $c=2000$, while the highest accuracy is observed at $\gamma=0.96$ and $c=1000$. Notably, the lowest ECE obtained is $1.33\%$ on the validation set, which is substantially lower than reported in Table \ref{tab:cifar100_calibration_methods}. While this comes at the cost of a slight decrease in accuracy, calibration may be prioritized for certain tasks or applications. Furthermore, by comparing $\gamma=1.0$ with the remainder of the values in Figures \ref{fig:ablation_acc} and \ref{fig:ablation_ece}, we note that the discounting is a valuable addition that appears to improve both accuracy and ECE. 

While DPS appears to offer low ECE and high accuracy for wide range of values of $\gamma$ and $c$, we note that if hyperparameters are chosen poorly (e.g., $\gamma=0.9$, $c=50.0$), performance deteriorates. This is likely due to underfitting caused by relying too much on predictions too early.

\section{Discussion}
\label{sec:discussion}
Overall, our results demonstrate that DPS can improve test set accuracy, ECE and NLL across a variety of datasets, architectures, and augmentation strategies compared to conventional training and related self-distillation methods. DPS does not seem to overfit to noise in the same way as conventional training and contrastive self-distillation methods, and achieves higher test set accuracy under label noise than other architecture-preserving self-distillation methods. Additionally, DPS+ yields state-of-the-art accuracy under label noise on CIFAR-10. 

\textbf{Computational cost and memory requirements.}  
The Bayesian update in Equation~\ref{eq:bayesian_update} requires first computing a per-example scalar weight from $A$ and then applying that weight to the two target tensors. For a mini-batch of size $B$ and $K$ classes, this is $\mathcal{O}(BK)$ and negligible compared with a forward-backward pass. Memory-wise, we store the target distributions $\mathbf{y}$ and per-example counts $A$. With float16 this requires $2NK$ bytes for $\mathbf{y}$ and $2N$ bytes for $A$. TE and PS-KD require the same amount of memory for storing the targets, whereas DLB requires a batch-wise buffer of $2BK$ bytes.

\newpage
\bibliographystyle{iclr2026_conference}  
\bibliography{iclr2026_conference}  

@String(NIPS= {Adv. Neural Inform. Process. Syst.})

@String(ACCV  = {ACCV})

@String(AAAI = {AAAI})

@String(NIPS  = {NeurIPS})

@article{hinton2015distilling,
  title={Distilling the knowledge in a neural network},
  author={Hinton, Geoffrey and Vinyals, Oriol and Dean, Jeff},
  journal={arXiv preprint arXiv:1503.02531},
  year={2015}
}

@inproceedings{szegedy2016rethinking,
  title={Rethinking the inception architecture for computer vision},
  author={Szegedy, Christian and Vanhoucke, Vincent and Ioffe, Sergey and Shlens, Jon and Wojna, Zbigniew},
  booktitle={Proceedings of the IEEE conference on computer vision and pattern recognition},
  pages={2818--2826},
  year={2016}
}

@article{laine2016temporal,
  title={Temporal ensembling for semi-supervised learning},
  author={Laine, Samuli and Aila, Timo},
  journal={arXiv preprint arXiv:1610.02242},
  year={2016}
}

@inproceedings{furlanello2018born,
  title={Born again neural networks},
  author={Furlanello, Tommaso and Lipton, Zachary and Tschannen, Michael and Itti, Laurent and Anandkumar, Anima},
  booktitle={International conference on machine learning},
  pages={1607--1616},
  year={2018},
  organization={PMLR}
}

@inproceedings{yang2019snapshot,
  title={Snapshot distillation: Teacher-student optimization in one generation},
  author={Yang, Chenglin and Xie, Lingxi and Su, Chi and Yuille, Alan L},
  booktitle={Proceedings of the IEEE/CVF Conference on Computer Vision and Pattern Recognition},
  pages={2859--2868},
  year={2019}
}

@inproceedings{zhang2019your,
  title={Be your own teacher: Improve the performance of convolutional neural networks via self distillation},
  author={Zhang, Linfeng and Song, Jiebo and Gao, Anni and Chen, Jingwei and Bao, Chenglong and Ma, Kaisheng},
  booktitle={Proceedings of the IEEE/CVF international conference on computer vision},
  pages={3713--3722},
  year={2019}
}

@inproceedings{yun2020regularizing,
  title={Regularizing class-wise predictions via self-knowledge distillation},
  author={Yun, Sukmin and Park, Jongjin and Lee, Kimin and Shin, Jinwoo},
  booktitle={Proceedings of the IEEE/CVF conference on computer vision and pattern recognition},
  pages={13876--13885},
  year={2020}
}

@article{tarvainen2017mean,
  title={Mean teachers are better role models: Weight-averaged consistency targets improve semi-supervised deep learning results},
  author={Tarvainen, Antti and Valpola, Harri},
  journal={Advances in neural information processing systems},
  volume={30},
  year={2017}
}

@inproceedings{kim2021self,
  title={Self-knowledge distillation with progressive refinement of targets},
  author={Kim, Kyungyul and Ji, ByeongMoon and Yoon, Doyoung and Hwang, Sangheum},
  booktitle={Proceedings of the IEEE/CVF international conference on computer vision},
  pages={6567--6576},
  year={2021}
}

@inproceedings{shen2022self,
  title={Self-distillation from the last mini-batch for consistency regularization},
  author={Shen, Yiqing and Xu, Liwu and Yang, Yuzhe and Li, Yaqian and Guo, Yandong},
  booktitle={Proceedings of the IEEE/CVF conference on computer vision and pattern recognition},
  pages={11943--11952},
  year={2022}
}

@misc{krizhevsky2009learning,
  title={Learning multiple layers of features from tiny images.(2009)},
  author={Krizhevsky, Alex and Hinton, Geoffrey and others},
  year={2009}
}

@article{russakovsky2015imagenet,
  title={Imagenet large scale visual recognition challenge},
  author={Russakovsky, Olga and Deng, Jia and Su, Hao and Krause, Jonathan and Satheesh, Sanjeev and Ma, Sean and Huang, Zhiheng and Karpathy, Andrej and Khosla, Aditya and Bernstein, Michael and others},
  journal={International journal of computer vision},
  volume={115},
  pages={211--252},
  year={2015},
  publisher={Springer}
}

@inproceedings{he2016deep,
  title={Deep residual learning for image recognition},
  author={He, Kaiming and Zhang, Xiangyu and Ren, Shaoqing and Sun, Jian},
  booktitle={Proceedings of the IEEE conference on computer vision and pattern recognition},
  pages={770--778},
  year={2016}
}

@inproceedings{huang2017densely,
  title={Densely connected convolutional networks},
  author={Huang, Gao and Liu, Zhuang and Van Der Maaten, Laurens and Weinberger, Kilian Q},
  booktitle={Proceedings of the IEEE conference on computer vision and pattern recognition},
  pages={4700--4708},
  year={2017}
}

@article{dosovitskiy2020image,
  title={An image is worth 16x16 words: Transformers for image recognition at scale},
  author={Dosovitskiy, Alexey and Beyer, Lucas and Kolesnikov, Alexander and Weissenborn, Dirk and Zhai, Xiaohua and Unterthiner, Thomas and Dehghani, Mostafa and Minderer, Matthias and Heigold, Georg and Gelly, Sylvain and others},
  journal={arXiv preprint arXiv:2010.11929},
  year={2020}
}

@misc{tinyimagenet,
  title        = {Tiny ImageNet Challenge},
  author       = {{Stanford CS231n}},
  year         = {2017},
  howpublished = {\url{http://cs231n.stanford.edu/tiny-imagenet-200.zip}},
  note         = {Accessed: 2025-02-20}
}

@misc{paszke2019pytorch,
      title={PyTorch: An Imperative Style, High-Performance Deep Learning Library}, 
      author={Adam Paszke and Sam Gross and Francisco Massa and Adam Lerer and James Bradbury and Gregory Chanan and Trevor Killeen and Zeming Lin and Natalia Gimelshein and Luca Antiga and Alban Desmaison and Andreas Köpf and Edward Yang and Zach DeVito and Martin Raison and Alykhan Tejani and Sasank Chilamkurthy and Benoit Steiner and Lu Fang and Junjie Bai and Soumith Chintala},
      year={2019},
      eprint={1912.01703},
      archivePrefix={arXiv},
      primaryClass={cs.LG},
      url={https://arxiv.org/abs/1912.01703}, 
}

@inproceedings{smith2017superconvergence,
  title={Super-convergence: Very fast training of neural networks using large learning rates},
  author={Smith, Leslie N and Topin, Nicholay},
  booktitle={Artificial intelligence and machine learning for multi-domain operations applications},
  volume={11006},
  pages={369--386},
  year={2019},
  organization={SPIE}
}

@inproceedings{yun2019cutmix,
  title={Cutmix: Regularization strategy to train strong classifiers with localizable features},
  author={Yun, Sangdoo and Han, Dongyoon and Oh, Seong Joon and Chun, Sanghyuk and Choe, Junsuk and Yoo, Youngjoon},
  booktitle={Proceedings of the IEEE/CVF international conference on computer vision},
  pages={6023--6032},
  year={2019}
}

@article{zhang2022self,
  title={Self-distillation: Towards efficient and compact neural networks},
  author={Zhang, Linfeng and Bao, Chenglong and Ma, Kaisheng},
  journal={IEEE Transactions on Pattern Analysis and Machine Intelligence},
  volume={44},
  number={8},
  pages={4388--4403},
  year={2021},
  publisher={IEEE}
}

@article{lan2018knowledge,
  title={Knowledge distillation by on-the-fly native ensemble},
  author={Zhu, Xiatian and Gong, Shaogang and others},
  journal={Advances in neural information processing systems},
  volume={31},
  year={2018}
}

@article{devries2017improved,
  title={Improved regularization of convolutional neural networks with cutout},
  author={DeVries, Terrance and Taylor, Graham W},
  journal={arXiv preprint arXiv:1708.04552},
  year={2017}
}

@inproceedings{guo2017calibration,
  title={On calibration of modern neural networks},
  author={Guo, Chuan and Pleiss, Geoff and Sun, Yu and Weinberger, Kilian Q},
  booktitle={International conference on machine learning},
  pages={1321--1330},
  year={2017},
  organization={PMLR}
}

@inproceedings{naeini2015obtaining,
  title={Obtaining well calibrated probabilities using bayesian binning},
  author={Naeini, Mahdi Pakdaman and Cooper, Gregory and Hauskrecht, Milos},
  booktitle={Proceedings of the AAAI conference on artificial intelligence},
  volume={29},
  number={1},
  year={2015}
}

@article{nakkiran2021deep,
  title={Deep double descent: Where bigger models and more data hurt},
  author={Nakkiran, Preetum and Kaplun, Gal and Bansal, Yamini and Yang, Tristan and Barak, Boaz and Sutskever, Ilya},
  journal={Journal of Statistical Mechanics: Theory and Experiment},
  volume={2021},
  number={12},
  pages={124003},
  year={2021},
  publisher={IOP Publishing}
}

@book{bishop2006pattern,
  title={Pattern recognition and machine learning},
  author={Bishop, Christopher M and Nasrabadi, Nasser M},
  volume={4},
  number={4},
  year={2006},
  publisher={Springer},
  note      = {See Section 2.2.1 on page 76.}
}

@inproceedings{feng2023ot,
  title={OT-Filter: An optimal transport filter for learning with noisy labels},
  author={Feng, Chuanwen and Ren, Yilong and Xie, Xike},
  booktitle={Proceedings of the IEEE/CVF Conference on Computer Vision and Pattern Recognition},
  pages={16164--16174},
  year={2023}
}

@misc{hinton2014dark,
  author       = {Geoffrey Hinton},
  title        = {Dark Knowledge},
  howpublished = {Distinguished Lecture Series, Toyota Technological Institute at Chicago},
  year         = {2014},
  month        = {October},
  day          = {2},
  note         = {https://www.ttic.edu/dls-2014-2015/},
}

@inproceedings{hebbalaguppe2024calibration,
  title={Calibration Transfer via Knowledge Distillation},
  author={Hebbalaguppe, R. et al.},
  booktitle={ACCV},
  year={2024}
}

@inbook{west2006chapter7,
  author    = {West, Mike and Harrison, Jeff},
  title     = {Bayesian Forecasting and Dynamic Models},
  chapter   = {6},
  pages     = {193--202},
  publisher = {Springer},
  year      = {2006}
}

@article{kingma2014adam,
  title={Adam: A method for stochastic optimization},
  author={Kingma, Diederik P},
  journal={arXiv preprint arXiv:1412.6980},
  year={2014}
}

@article{hendrycks2019benchmarking,
  title={Benchmarking neural network robustness to common corruptions and perturbations},
  author={Hendrycks, Dan and Dietterich, Thomas},
  journal={arXiv preprint arXiv:1903.12261},
  year={2019}
}

@article{muller2019does,
  title={When does label smoothing help?},
  author={M{\"u}ller, Rafael and Kornblith, Simon and Hinton, Geoffrey E},
  journal={Advances in neural information processing systems},
  volume={32},
  year={2019}
}

@inproceedings{tanaka2018joint,
  title={Joint optimization framework for learning with noisy labels},
  author={Tanaka, Daiki and Ikami, Daiki and Yamasaki, Toshihiko and Aizawa, Kiyoharu},
  booktitle={Proceedings of the IEEE conference on computer vision and pattern recognition},
  pages={5552--5560},
  year={2018}
}

@inproceedings{wang2019symmetric,
  title={Symmetric cross entropy for robust learning with noisy labels},
  author={Wang, Yisen and Ma, Xingjun and Chen, Zaiyi and Luo, Yuan and Yi, Jinfeng and Bailey, James},
  booktitle={Proceedings of the IEEE/CVF international conference on computer vision},
  pages={322--330},
  year={2019}
}

@article{zhang2017mixup,
  title={mixup: Beyond empirical risk minimization},
  author={Zhang, Hongyi and Cisse, Moustapha and Dauphin, Yann N and Lopez-Paz, David},
  journal={arXiv preprint arXiv:1710.09412},
  year={2017}
}

@article{chang2023csot,
  title={Csot: Curriculum and structure-aware optimal transport for learning with noisy labels},
  author={Chang, Wanxing and Shi, Ye and Wang, Jingya},
  journal={Advances in Neural Information Processing Systems},
  volume={36},
  pages={8528--8541},
  year={2023}
}

@inproceedings{xu2025revisiting,
  title={Revisiting Interpolation for Noisy Label Correction},
  author={Xu, Yuanzhuo and Niu, Xiaoguang and Yang, Jie and Su, Ruiyi and Zhang, Jian and Liu, Shubo and Drew, Steve},
  booktitle={Proceedings of the AAAI Conference on Artificial Intelligence},
  volume={39},
  number={20},
  pages={21833--21841},
  year={2025}
}

@inproceedings{he2016identity,
  title={Identity mappings in deep residual networks},
  author={He, Kaiming and Zhang, Xiangyu and Ren, Shaoqing and Sun, Jian},
  booktitle={Computer Vision--ECCV 2016: 14th European Conference, Amsterdam, The Netherlands, October 11--14, 2016, Proceedings, Part IV 14},
  pages={630--645},
  year={2016},
  organization={Springer}
}

@article{loshchilov2016sgdr,
  title={Sgdr: Stochastic gradient descent with warm restarts},
  author={Loshchilov, Ilya and Hutter, Frank},
  journal={arXiv preprint arXiv:1608.03983},
  year={2016}
}

@inproceedings{zhong2020random,
  title={Random erasing data augmentation},
  author={Zhong, Zhun and Zheng, Liang and Kang, Guoliang and Li, Shaozi and Yang, Yi},
  booktitle={Proceedings of the AAAI conference on artificial intelligence},
  volume={34},
  number={07},
  pages={13001--13008},
  year={2020}
}

@article{liu2023chimera,
  title={ChiMera: Learning with noisy labels by contrasting mixed-up augmentations},
  author={Liu, Zixuan and Zhang, Xin and He, Junjun and Fu, Dan and Samaras, Dimitris and Tan, Robby and Wang, Xiao and Wang, Sheng},
  journal={arXiv preprint arXiv:2310.05183},
  year={2023}
}

@article{zhang2024bpt,
  title={BPT-PLR: A balanced partitioning and training framework with pseudo-label relaxed contrastive loss for noisy label learning},
  author={Zhang, Qian and Jin, Ge and Zhu, Yi and Wei, Hongjian and Chen, Qiu},
  journal={Entropy},
  volume={26},
  number={7},
  pages={589},
  year={2024},
  publisher={MDPI}
}

@inproceedings{blundell2015weight,
  title={Weight uncertainty in neural network},
  author={Blundell, Charles and Cornebise, Julien and Kavukcuoglu, Koray and Wierstra, Daan},
  booktitle={International conference on machine learning},
  pages={1613--1622},
  year={2015},
  organization={PMLR}
}

@inproceedings{gal2016dropout,
  title={Dropout as a bayesian approximation: Representing model uncertainty in deep learning},
  author={Gal, Yarin and Ghahramani, Zoubin},
  booktitle={international conference on machine learning},
  pages={1050--1059},
  year={2016},
  organization={PMLR}
}

@article{lakshminarayanan2017simple,
  title={Simple and scalable predictive uncertainty estimation using deep ensembles},
  author={Lakshminarayanan, Balaji and Pritzel, Alexander and Blundell, Charles},
  journal={Advances in neural information processing systems},
  volume={30},
  year={2017}
}

@inproceedings{moon2020confidence,
  title={Confidence-aware learning for deep neural networks},
  author={Moon, Jooyoung and Kim, Jihyo and Shin, Younghak and Hwang, Sangheum},
  booktitle={international conference on machine learning},
  pages={7034--7044},
  year={2020},
  organization={PMLR}
}

@inproceedings{hebbalaguppe2022stitch,
  title={A stitch in time saves nine: A train-time regularizing loss for improved neural network calibration},
  author={Hebbalaguppe, Ramya and Prakash, Jatin and Madan, Neelabh and Arora, Chetan},
  booktitle={Proceedings of the IEEE/CVF Conference on Computer Vision and Pattern Recognition},
  pages={16081--16090},
  year={2022}
}

@article{ghosh2022adafocal,
  title={Adafocal: Calibration-aware adaptive focal loss},
  author={Ghosh, Arindam and Schaaf, Thomas and Gormley, Matthew},
  journal={Advances in Neural Information Processing Systems},
  volume={35},
  pages={1583--1595},
  year={2022}
}

@inproceedings{cheng2022calibrating,
  title={Calibrating deep neural networks by pairwise constraints},
  author={Cheng, Jiacheng and Vasconcelos, Nuno},
  booktitle={Proceedings of the IEEE/CVF Conference on Computer Vision and Pattern Recognition},
  pages={13709--13718},
  year={2022}
}

@inproceedings{liu2022devil,
  title={The devil is in the margin: Margin-based label smoothing for network calibration},
  author={Liu, Bingyuan and Ben Ayed, Ismail and Galdran, Adrian and Dolz, Jose},
  booktitle={Proceedings of the IEEE/CVF Conference on Computer Vision and Pattern Recognition},
  pages={80--88},
  year={2022}
}

@inproceedings{park2023acls,
  title={Acls: Adaptive and conditional label smoothing for network calibration},
  author={Park, Hyekang and Noh, Jongyoun and Oh, Youngmin and Baek, Donghyeon and Ham, Bumsub},
  booktitle={Proceedings of the IEEE/CVF International Conference on Computer Vision},
  pages={3936--3945},
  year={2023}
}

@inproceedings{netzer2011reading,
  title={Reading digits in natural images with unsupervised feature learning},
  author={Netzer, Yuval and Wang, Tao and Coates, Adam and Bissacco, Alessandro and Wu, Baolin and Ng, Andrew Y and others},
  booktitle={NIPS workshop on deep learning and unsupervised feature learning},
  volume={2011},
  number={5},
  pages={7},
  year={2011},
  organization={Granada}
}

@article{singhal1988training,
  title={Training multilayer perceptrons with the extended Kalman algorithm},
  author={Singhal, Sharad and Wu, Lance},
  journal={Advances in neural information processing systems},
  volume={1},
  year={1988}
}

@article{mandt2017stochastic,
  title={Stochastic gradient descent as approximate bayesian inference},
  author={Mandt, Stephan and Hoffman, Matthew D and Blei, David M},
  journal={Journal of Machine Learning Research},
  volume={18},
  number={134},
  pages={1--35},
  year={2017}
}

@book{sarkka2023bayesian,
  title={Bayesian filtering and smoothing},
  author={S{\"a}rkk{\"a}, Simo and Svensson, Lennart},
  volume={17},
  year={2023},
  publisher={Cambridge university press}
}

@inproceedings{muller2021trivialaugment,
  title={Trivialaugment: Tuning-free yet state-of-the-art data augmentation},
  author={M{\"u}ller, Samuel G and Hutter, Frank},
  booktitle={Proceedings of the IEEE/CVF international conference on computer vision},
  pages={774--782},
  year={2021}
}

\clearpage
\appendix
\section{Appendix}
\subsection{Experimental Details}
Unless otherwise stated, all models trained on CIFAR and TinyImageNet datasets use the Adam optimizer \citep{kingma2014adam} with a batch size of $256$, a maximum learning rate of $0.01$ scheduled via the 1cycle policy \citep{smith2017superconvergence}, and basic augmentations (random cropping and horizontal flipping) for 200 epochs (40 for ViTs, with a maximum learning rate of 5e-5). For ImageNet, we train with SGD for $200$ epochs using a per-GPU batch size of $128$ on 8 GPUs, a learning rate of $0.5$ (scheduled via cosine annealing with 5 epochs of warm-up), a weight decay of $1e-4$, and random resized crops of $224$ pixels combined with horizontal flipping. 

\textbf{Architectures.} Because of the small size of the images, the ResNet and DenseNet networks have been modified to include a $3 \times 3$ convolution instead of the usual $7 \times 7$ convolution for all datasets except ImageNet. The ViT-B \citep{dosovitskiy2020image} model has been pretrained on ImageNet for all experiments and is available in Pytorch \citep{paszke2019pytorch}.

\textbf{Method-specific hyperparameters.} For the temporal ensemble, we set the momentum parameter $\alpha=0.6$, gradually ramping up the distillation loss over the first 100 epochs (45 for ImageNet) and anneal Adam’s $\beta_1$ to zero during the final 50 epochs (25 for ImageNet). For DLB, we follow \citet{shen2022self} and use a temperature of $3$ and set $\alpha = 1.0$, but train for only 100 epochs to offset the doubled mini-batch size. For PS-KD, we let $\alpha_T = 0.8$ ($\alpha_T = 0.3$ for ImageNet). For the knowledge distillation of the ensembles, we use a temperature of 3.

\textbf{Label Noise.} Inspired by methods in semi-supervised learning, we construct DPS+ by combining DPS with a contrastive loss term. We do this by utilizing a strong and a weak set of augmentations, where the weak set is used for DPS, and the strong is used for the contrastive term. We define the contrastive loss as
\begin{equation}
        \mathcal{L}_c = \frac{1}{m} \sum_{j=1}^m KL(f(T_j(\mathbf{x}_i)),\hat{\mathbf{y}}^t),
\end{equation}
where $m$ is the number of strongly augmented views, and $T_j$ is the corresponding transform, which is added to the supervised loss like $\mathcal{L}+\lambda_c\mathcal{L}_c$. For the strong set of augmentations, we utilize TrivialAugment \citep{muller2021trivialaugment}, in combination with CutMix \citep{yun2019cutmix} and Random Erasing \citep{zhong2020random}. For DPS+, we set $m=2$ for all noise levels and schedule the learning rate using cosine annealing \citep{loshchilov2016sgdr}.

Because DPS may excessively smooth the label distributions when labels are noisy, we introduce a sharpening parameter $\tau$ into the loss 
\begin{equation}
    \mathcal{L} \gets \frac{1}{|B|} \sum_{i \in B} \ell\left(\hat{\mathbf{y}}_i^{t}, \frac{\left(\mathbf{y}_i^{t-1}\right)^{1/\tau}}{\|\left(\mathbf{y}_i^{t-1}\right)^{1/\tau}\|_1}\right),
\end{equation}
for both DPS and DPS+, where $|B|$ is the batch size. In the absence of noise, we use $\tau = 1$ to avoid promoting overconfident predictions, while we set $\tau=0.8$ for all experiments with label noise. For DPS, we set $\gamma = 0.85$ and $c=2000$ when injecting symmetric noise, and $\gamma = 0.9$ and $c=1000$ for asymmetric noise. For DPS+, we set $\gamma = 0.95$ and $c=1000$ for all experiments, and set $\lambda_c=2$ under asymmetric noise. For DPS+ under symmetric noise we set $\lambda_c=4$, $\lambda_c=7$ and $\lambda_c=14$ for noise levels 20\%, 50\% and 80\%, respectively.

\subsection{Additional Experiments}

\begin{figure*}[t]
    \centering
    \begin{minipage}[t]{0.48\textwidth}
        \centering
        \begin{subfigure}[t]{0.49\linewidth}
            \centering
            \includegraphics[trim={0.5cm 0.7cm 0.2cm 0.4cm}, clip, width=\linewidth]{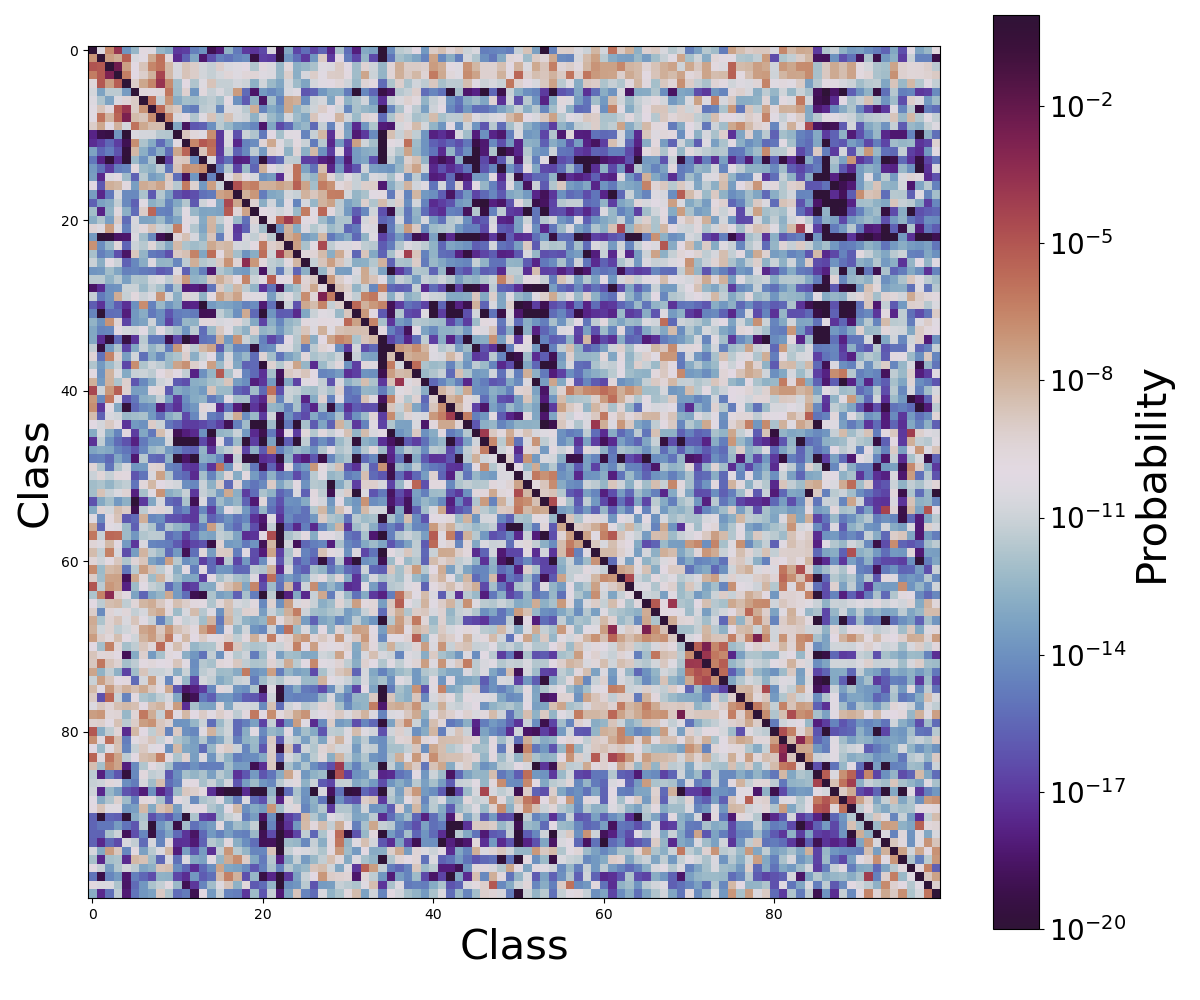}
            \caption{Conventional}
        \end{subfigure}
        \hfill
        \begin{subfigure}[t]{0.49\linewidth}
            \centering
            \includegraphics[trim={0.5cm 0.7cm 0.2cm 0.4cm}, clip, width=\linewidth]{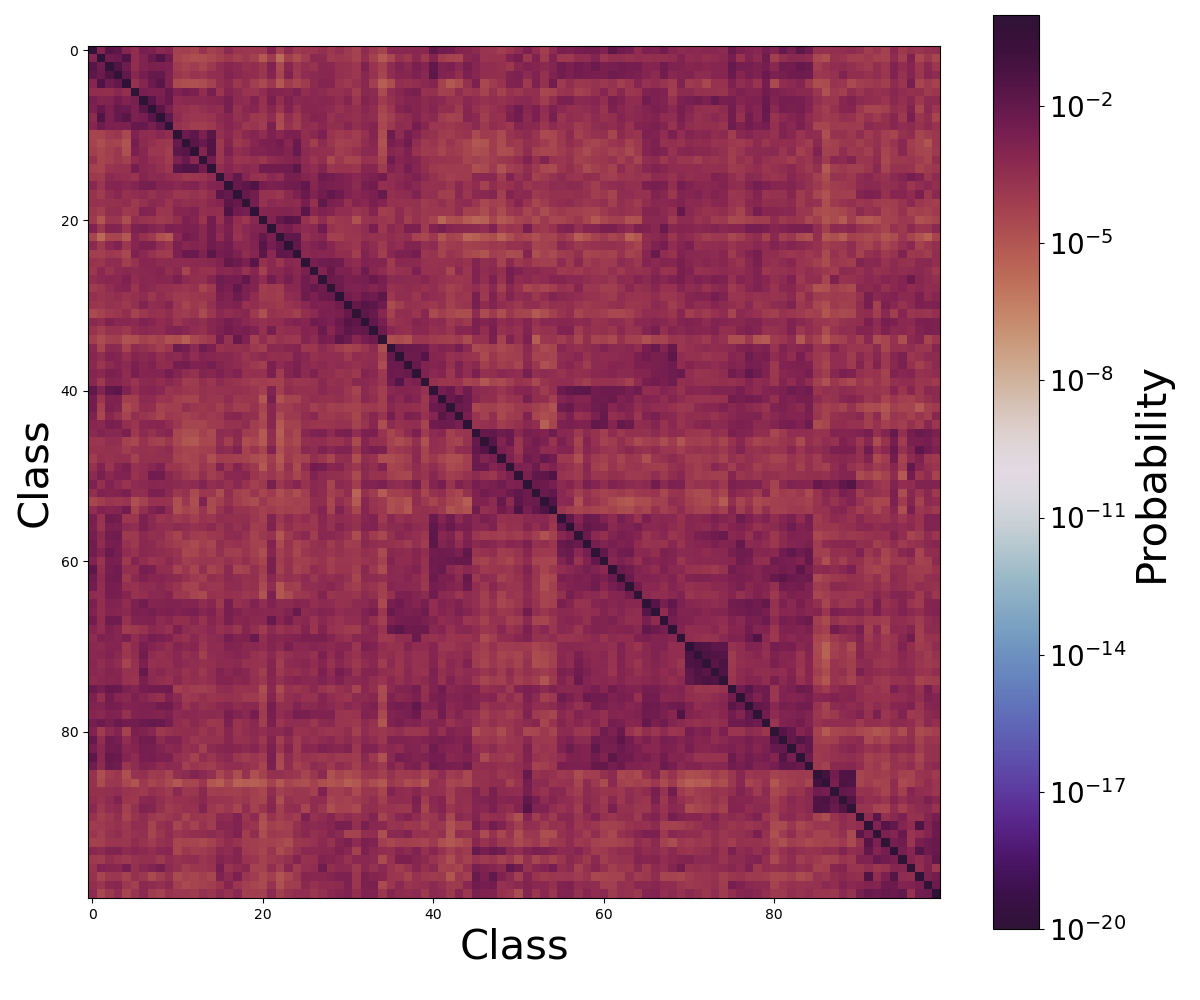}
            \caption{DPS}
        \end{subfigure}
        
        \caption{\textbf{Inter-class component $\bm\mu$ of dark knowledge.} Semantical patterns emerge between classes, accentuated by DPS (ResNet-50, CIFAR-100).}
        \label{fig:S_cifar100}
    \end{minipage}
    \hfill 
    \begin{minipage}[t]{0.48\textwidth}
        \centering
        \begin{subfigure}[t]{0.49\linewidth}
            \centering
            \includegraphics[width=\linewidth]{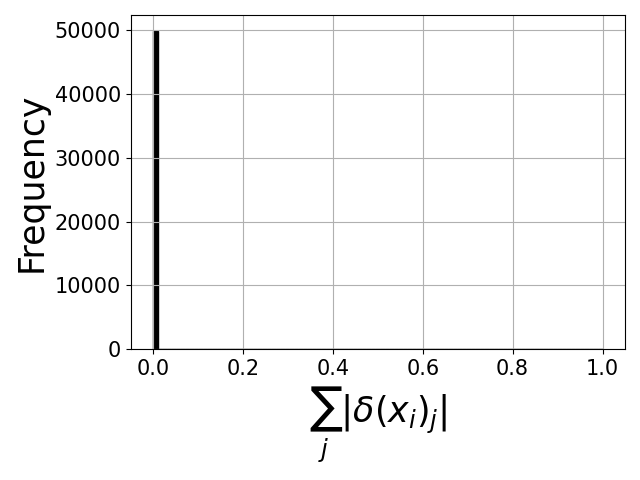}
            \caption{Conventional}
        \end{subfigure}
        \hfill
        \begin{subfigure}[t]{0.49\linewidth}
            \centering
            \includegraphics[width=\linewidth]{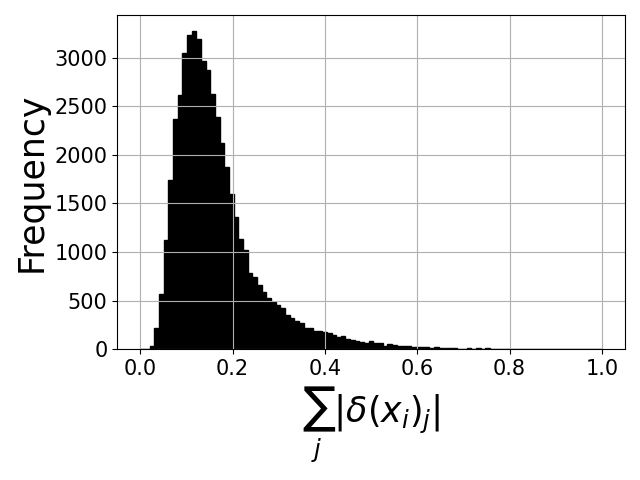}
            \caption{DPS}
        \end{subfigure}
        
        \caption{\textbf{Total sample-wise deviation $\delta(\mathbf{x}_i)$.} DPS promotes learning of sample-specific information (ResNet-50, CIFAR-100).}
        \label{fig:delta_cifar100}
    \end{minipage}
\end{figure*}

\begin{figure*}[t]
    \centering
    \begin{minipage}[t]{0.48\textwidth} 
        \centering
        \begin{subfigure}[t]{0.49\linewidth} 
            \centering
            \includegraphics[trim={0.5cm 0.7cm 0.2cm 0.4cm}, clip, width=\linewidth]{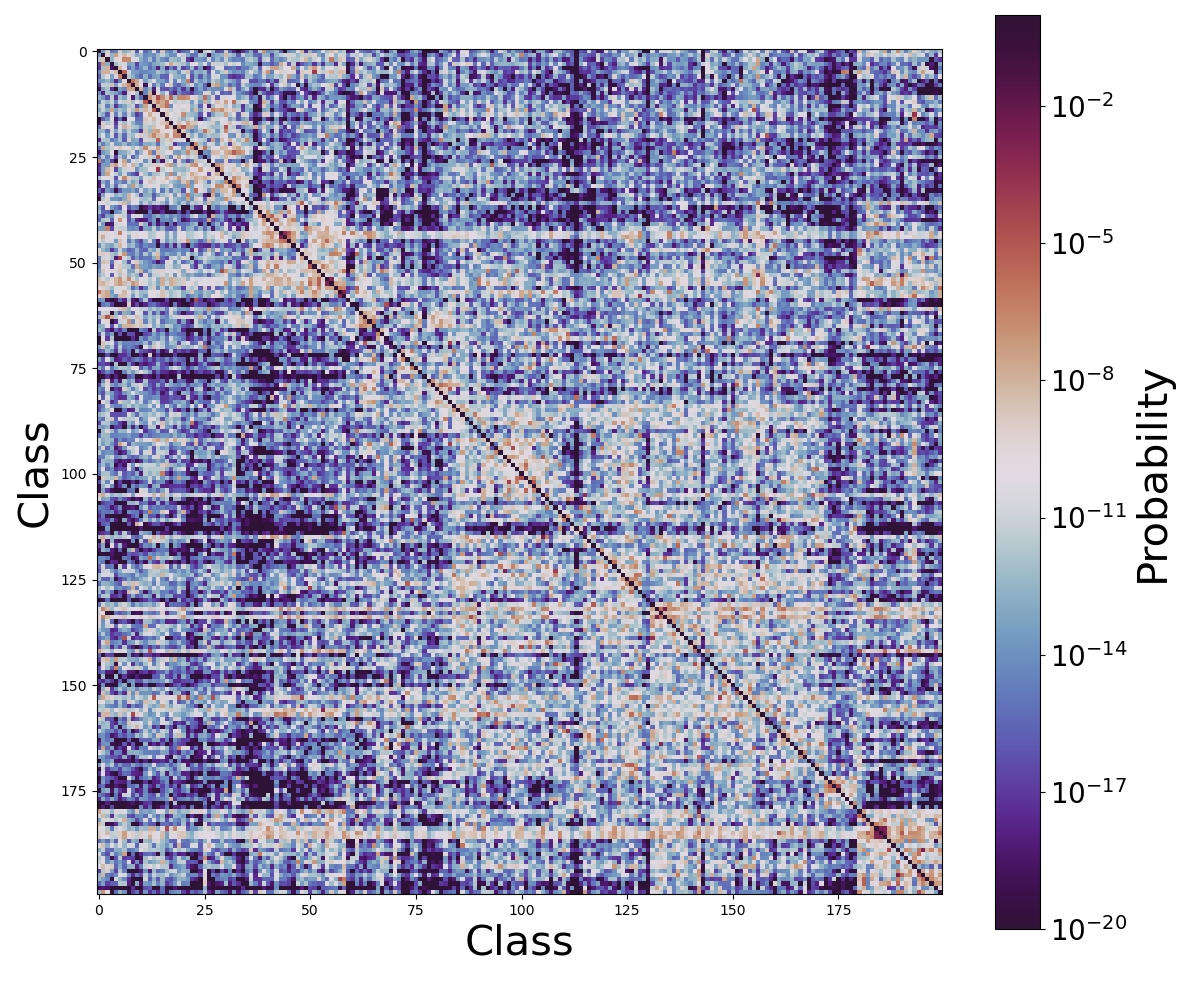}
            \caption{Conventional}
        \end{subfigure}
        \hfill
        \begin{subfigure}[t]{0.49\linewidth} 
            \centering
            \includegraphics[trim={0.5cm 0.7cm 0.2cm 0.4cm}, clip, width=\linewidth]{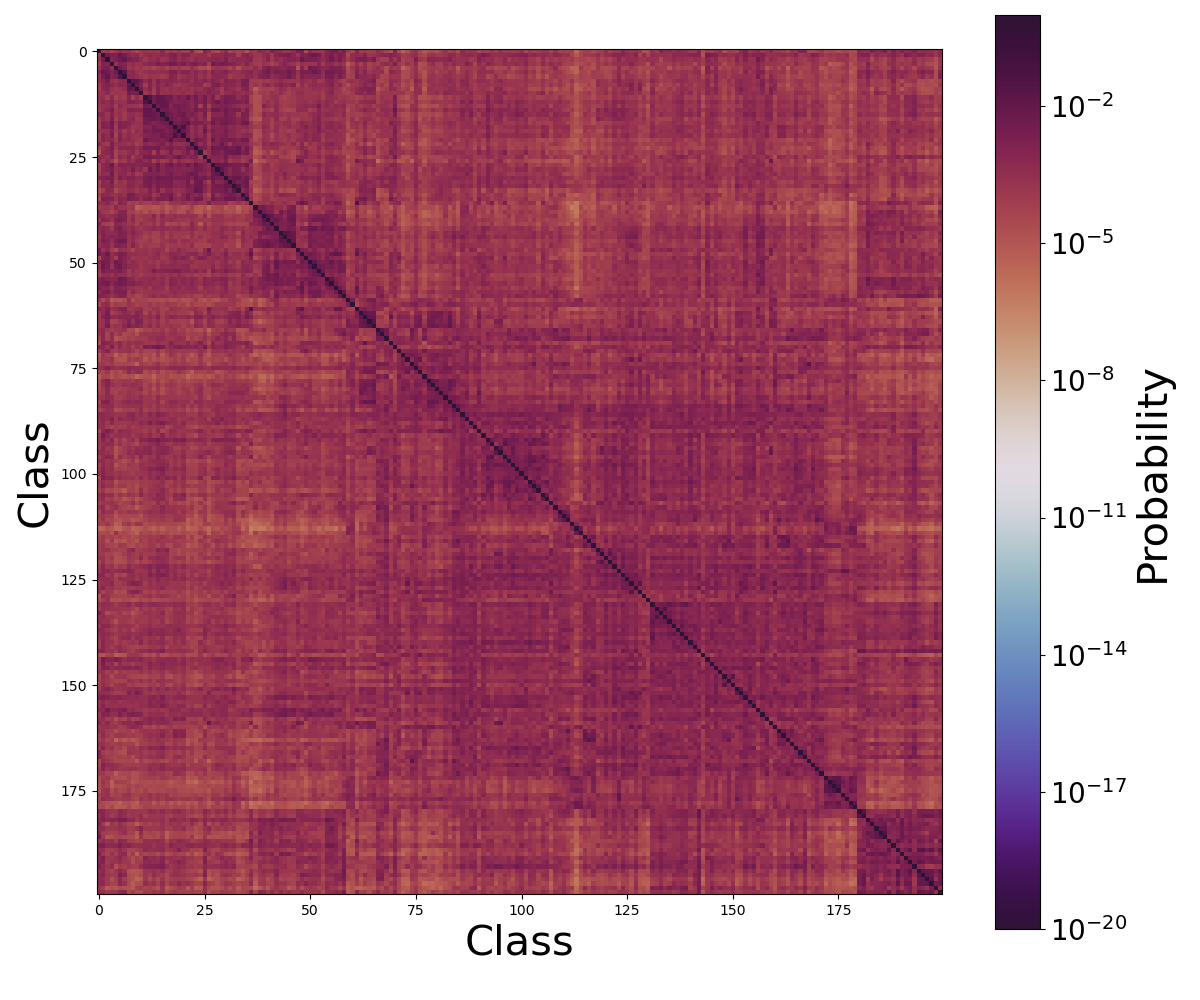}
            \caption{DPS}
        \end{subfigure}
        
        \caption{\textbf{Inter-class component $\bm\mu$ of dark knowledge.} Semantical patterns emerge between classes, accentuated by DPS (ResNet-101, Tiny ImageNet).}
        \label{fig:S_tinyimagenet}
    \end{minipage}
    \hfill 
    \begin{minipage}[t]{0.48\textwidth} 
        \centering
        \begin{subfigure}[t]{0.49\linewidth} 
            \centering
            \includegraphics[width=\linewidth]{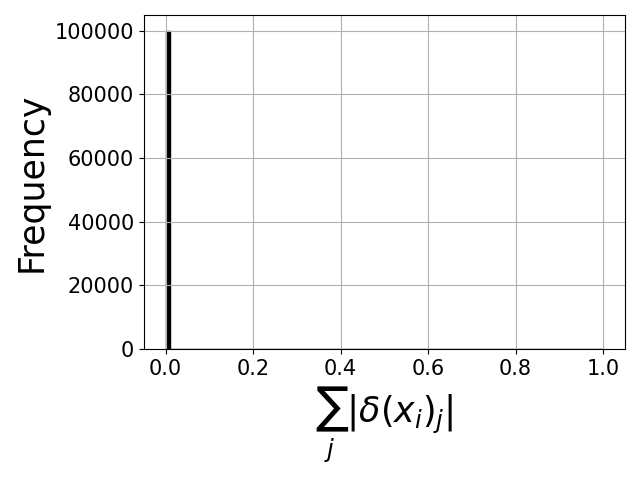}
            \caption{Conventional}
        \end{subfigure}
        \hfill
        \begin{subfigure}[t]{0.49\linewidth} 
            \centering
            \includegraphics[width=\linewidth]{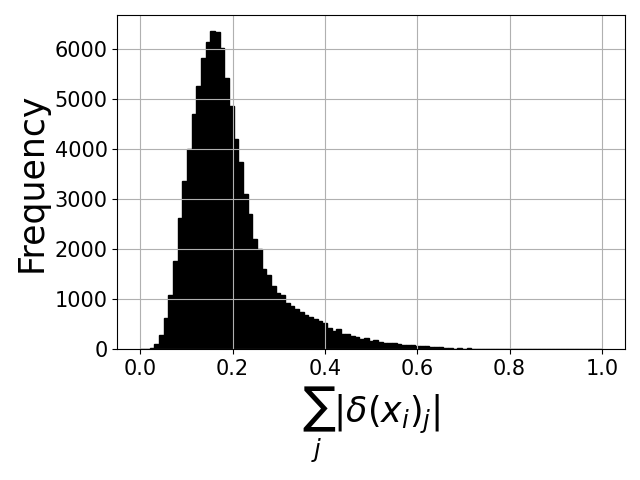}
            \caption{DPS}
        \end{subfigure}
        
        \caption{\textbf{Total sample-wise deviation $\delta(\mathbf{x}_i)$ of dark knowledge.} DPS promotes learning of sample-specific information (ResNet-101, Tiny ImageNet).}
        \label{fig:delta_tinyimagenet}
    \end{minipage}
\end{figure*}

\textbf{Analyzing dark knowledge.} We include visualizations of the inter-class distributions for CIFAR-100 and Tiny ImageNet in Figures \ref{fig:S_cifar100} and \ref{fig:S_tinyimagenet}, with their corresponding sample-wise deviations plotted in Figures \ref{fig:delta_cifar100} and \ref{fig:delta_tinyimagenet}, respectively. We observe similar patterns as for ResNet-18 in  Figures~\ref{fig:S_cifar10_conv}-\subref{fig:S_cifar10_dps}, but on a larger scale with increased sample-wise deviations.

\begin{figure}[ht]
    \centering    
    \begin{subfigure}{0.326\columnwidth}
        \centering
        \includegraphics[width=0.8\linewidth]{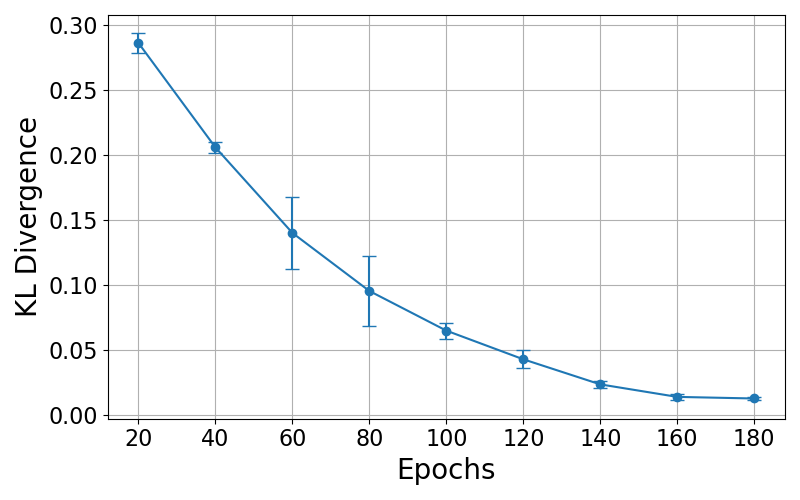}
        \caption{ResNet-18 on CIFAR-10}
    \end{subfigure}
  \hfill
    \begin{subfigure}{0.326\columnwidth}
        \centering
        \includegraphics[width=0.8\linewidth]{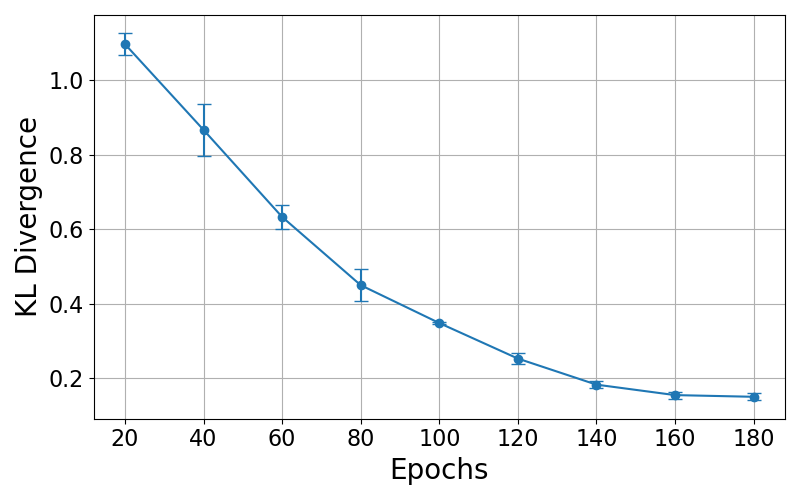}
        \caption{ResNet-50 on CIFAR-100}
    \end{subfigure}
    \hfill
    \begin{subfigure}{0.326\columnwidth}
        \centering
        \includegraphics[width=0.8\linewidth]{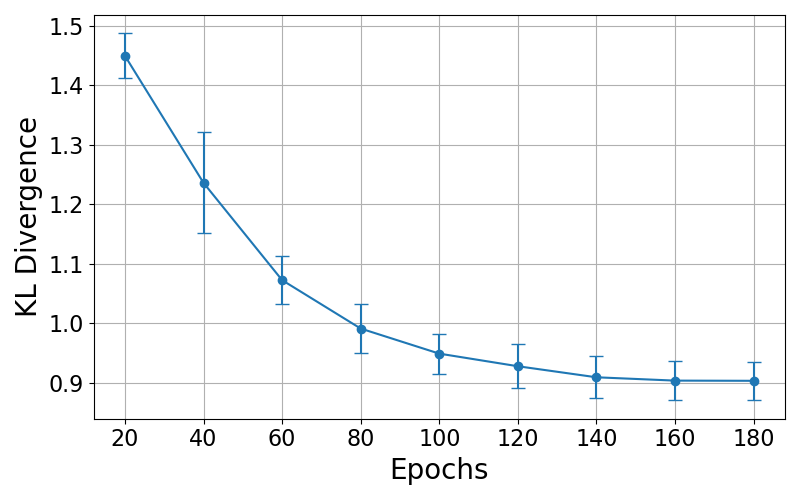}
        \caption{ResNet-101 on Tiny ImageNet}
    \end{subfigure}
    
    \caption{\textbf{Evolution of the average Temperature-Adjusted KL Divergence of predictions between the current and final model.} Dark knowledge emerges gradually during training.}
    \label{fig:dk}
\end{figure}
\begin{figure}[ht!]
  \centering
    \begin{subfigure}[b]{0.326\columnwidth}
    \centering
    \includegraphics[width=0.8\linewidth]{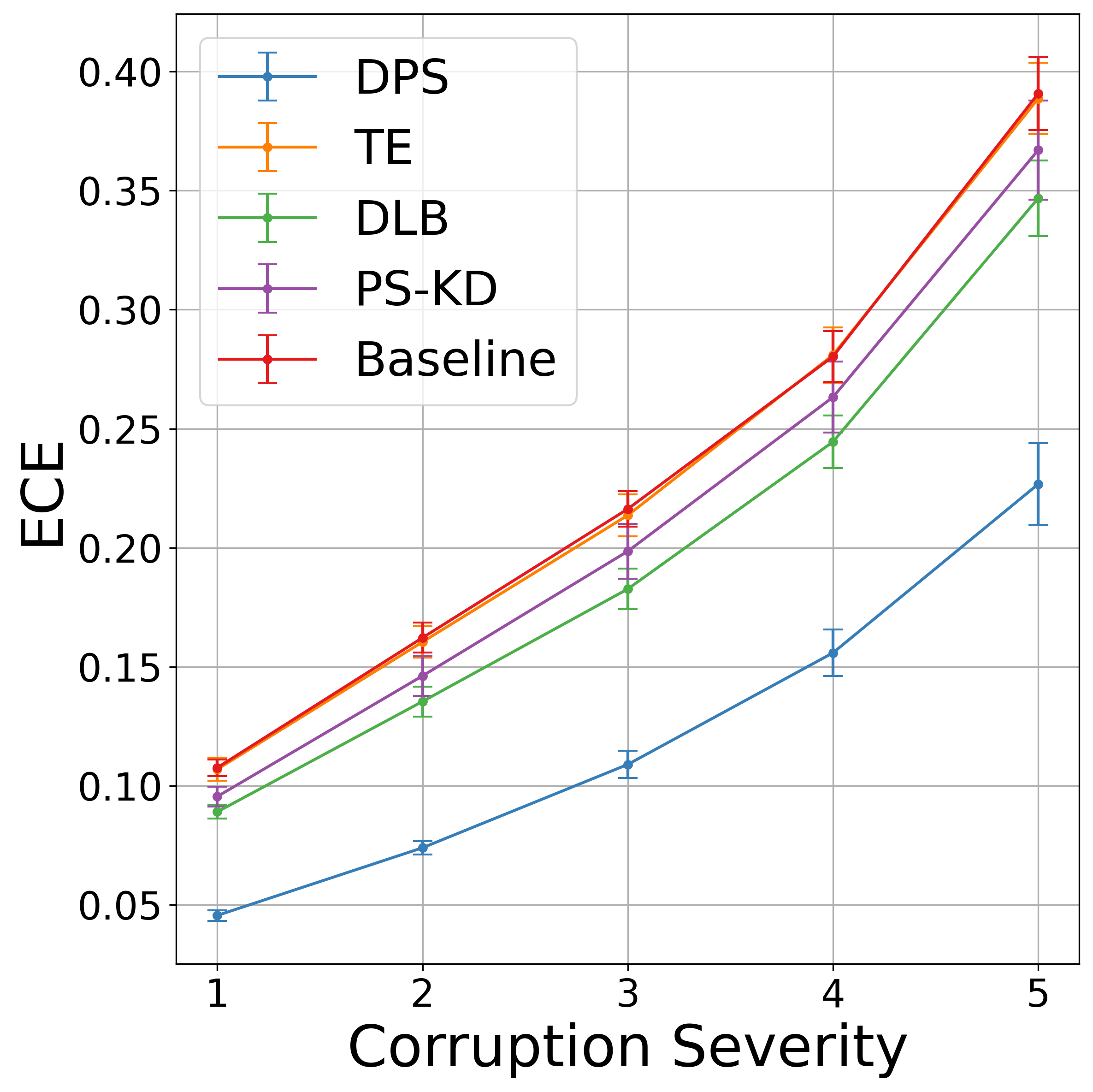}
    \caption{ResNet-50}
  \end{subfigure}
  \hfill
  \begin{subfigure}[b]{0.326\columnwidth}
    \centering
    \includegraphics[width=0.8\linewidth]{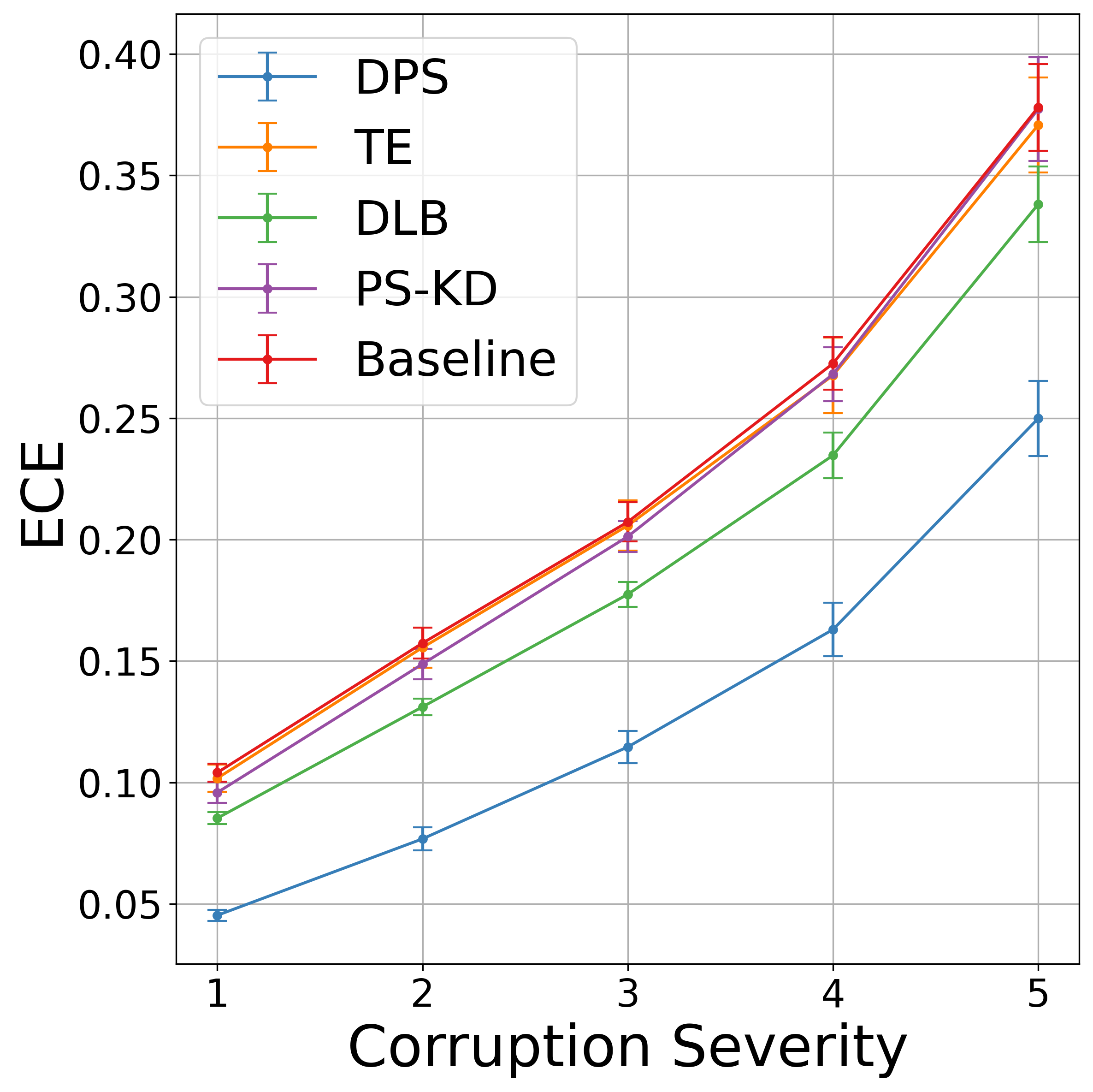}
    \caption{DenseNet-169}
  \end{subfigure}
  \hfill
  \begin{subfigure}[b]{0.326\columnwidth}
    \centering
    \includegraphics[width=0.8\linewidth]{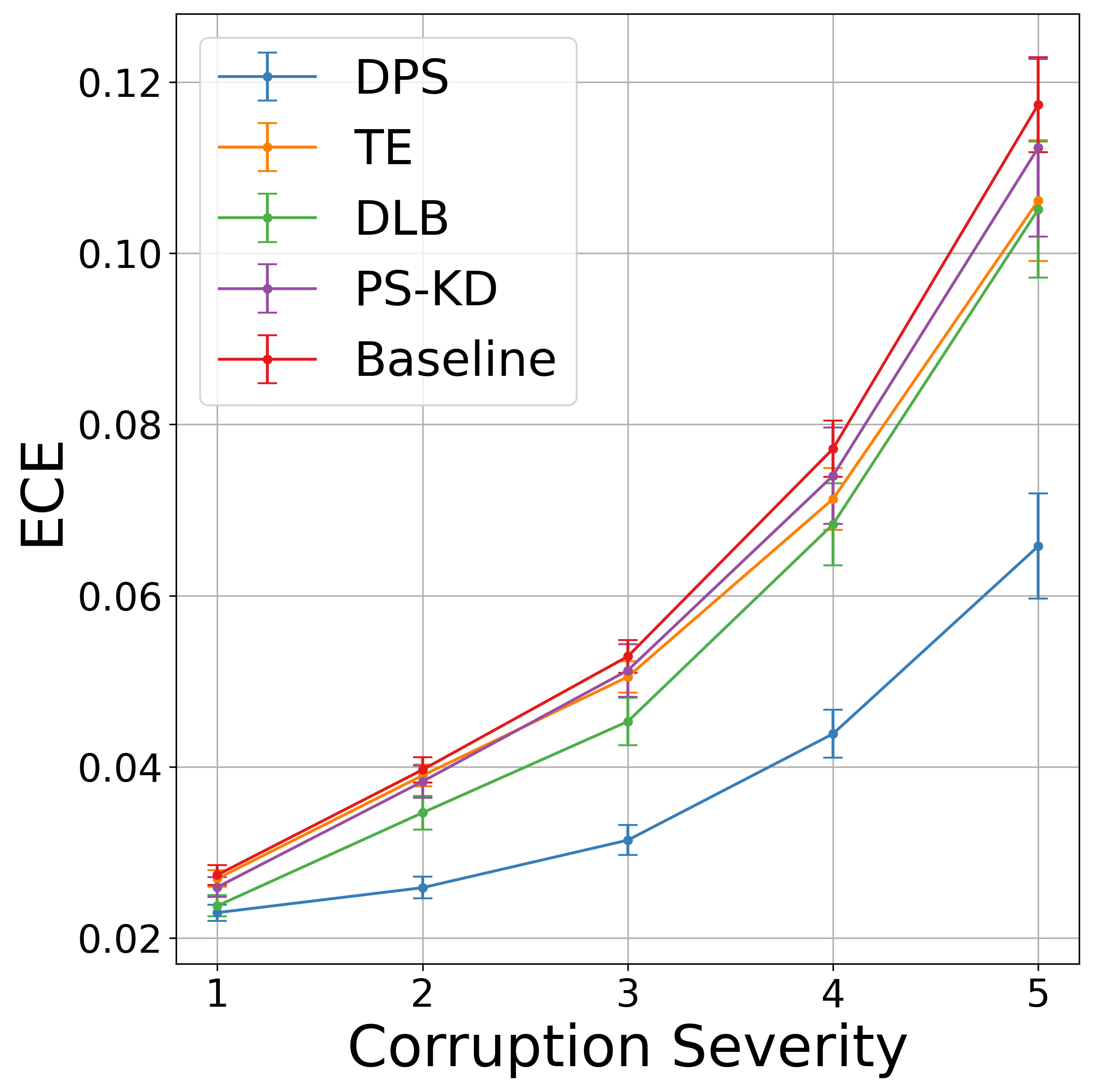}
    \caption{ViT-B}
  \end{subfigure}

  \caption{\textbf{ECE under increasing corruptions.} The models are trained using standard training (baseline), related methods \citep{laine2016temporal,shen2022self,kim2021self}, and the proposed method (DPS), on CIFAR-10 and evaluated on CIFAR-10-C.} 
  \label{fig:calibration_cifar10c}
\end{figure}

\begin{table*}[ht]
\centering
\caption{\textbf{Out-of-Distribution Detection Performance (AUROC).} Models trained on in-distribution datasets (CIFAR-10, CIFAR-100) are evaluated against SVHN as the out-of-distribution dataset. The models are trained using standard training (baseline), related methods \citep{laine2016temporal,shen2022self,kim2021self}, and the proposed method (DPS). The best results are highlighted in bold.}
\footnotesize
\setlength{\tabcolsep}{3pt}
\begin{tabular}{@{}llccccc@{}}
\toprule
ID Dataset & Model & Baseline (\%) & TE (\%) & DLB (\%) & PS-KD (\%) & DPS (\%) \\ \midrule
\multirow{3}{*}{CIFAR-10} 
 & ResNet-18    & $90.67{\scriptstyle\pm3.42}$ & $92.90{\scriptstyle\pm0.66}$ & $91.25{\scriptstyle\pm0.57}$ & $92.36{\scriptstyle\pm0.48}$ & $\mathbf{94.84}\,{\scriptscriptstyle\pm0.83}$ \\
 & DenseNet-121 & $90.96{\scriptstyle\pm4.22}$ & $\mathbf{94.05}\,{\scriptscriptstyle\pm0.98}$ & $91.19{\scriptstyle\pm1.20}$ & $91.11{\scriptstyle\pm0.64}$ & $93.65{\scriptstyle\pm0.16}$ \\
 & ViT-B/16     & $98.21{\scriptstyle\pm0.29}$ & $\mathbf{98.35}\,{\scriptscriptstyle\pm0.56}$ & $98.15{\scriptstyle\pm0.22}$ & $98.28{\scriptstyle\pm0.50}$ & $98.03{\scriptstyle\pm0.21}$ \\ \midrule
\multirow{3}{*}{CIFAR-100} 
 & ResNet-50    & $74.35{\scriptstyle\pm2.23}$ & $74.68{\scriptstyle\pm1.53}$ & $\mathbf{82.51}\,{\scriptscriptstyle\pm0.61}$ & $76.66{\scriptstyle\pm2.40}$ & $78.19{\scriptstyle\pm1.19}$ \\
 & DenseNet-169 & $77.92{\scriptstyle\pm0.57}$ & $68.97{\scriptstyle\pm6.85}$ & $80.55{\scriptstyle\pm3.56}$ & $73.95{\scriptstyle\pm2.06}$ & $\mathbf{82.41}\,{\scriptscriptstyle\pm1.79}$ \\
 & ViT-B/16     & $88.42{\scriptstyle\pm1.19}$ & $89.12{\scriptstyle\pm0.64}$ & $89.52{\scriptstyle\pm1.16}$ & $88.82{\scriptstyle\pm0.39}$ & $\mathbf{90.71}\,{\scriptscriptstyle\pm0.50}$ \\ \bottomrule
\end{tabular}
\label{tab:ood_auroc}
\end{table*}

\begin{table}[ht]
  \centering
  \caption{\textbf{Calibration Performance for WideResNet-40-1 on CIFAR-100.} Result for all methods (excl. DPS and baseline) are from \citep{hebbalaguppe2024calibration}. The best results are highlighted in bold.}
  \label{tab:calibration_wrn}
  \footnotesize
  \setlength{\tabcolsep}{3pt}
  \begin{tabular}{@{} l r r r r @{}}
    \toprule
    Method & Accuracy (\%) & ECE (\%) & SCE (\%) & ACE (\%) \\
    \midrule
    Baseline (NLL) & $70.04$ & $11.16$ & $0.30$ & $11.19$ \\
    LS & $70.07$ & $1.30$ & $0.21$ & $1.49$ \\
    TS & $70.04$ & $2.57$ & $0.19$ & $2.50$ \\
    MMCE & $69.69$ & $7.34$ & $0.25$ & $7.37$ \\
    MixUp & $72.04$ & $2.57$ & $0.21$ & $2.52$ \\
    CRL & $65.80$ & $13.91$ & $0.37$ & $13.91$ \\
    PS-KD & $72.56$ & $3.73$ & $0.20$ & $3.72$ \\
    MDCA & $68.51$ & $1.35$ & $0.21$ & $1.34$ \\
    AdaFocal & $67.36$ & $2.10$ & $0.21$ & $1.97$ \\
    CPC & $69.99$ & $7.61$ & $0.23$ & $7.55$ \\
    MBLS & $69.97$ & $5.37$ & $0.22$ & $5.37$ \\
    ACLS & $69.92$ & $7.00$ & $0.23$ & $6.99$ \\
    \midrule
    KD  & $69.60$ & $15.18$ & $0.37$ & $15.18$ \\
    KD + MixUp & $72.48$ & $1.21$ & $0.20$ & $1.17$ \\
    KD + AdaFocal & $71.70$ & $1.19$ & $0.19$ & $1.34$ \\
    KD + CPC & $70.00$ & $9.02$ & $0.26$ & $9.01$ \\
    KD + MDCA & $71.07$ & $0.98$ & $0.20$ & $1.10$ \\
    KD + MMCE & $72.08$ & $2.02$ & $0.19$ & $1.95$ \\
    \midrule
    DPS ($\gamma=0.97, c=2000$) & $72.34{\scriptstyle\pm0.26}$ & $\mathbf{0.85}\,{\scriptscriptstyle\pm0.08}$ & $\mathbf{0.18}\,{\scriptscriptstyle\pm0.00}$ & $\mathbf{0.87}\,{\scriptscriptstyle\pm0.19}$\\
    DPS ($\gamma=0.96, c=3000$) & $72.62\,{\scriptscriptstyle\pm0.20}$ & $3.28{\scriptstyle\pm0.04}$ & $0.19{\scriptstyle\pm0.00}$ & $3.19{\scriptstyle\pm0.12}$\\
    DPS ($\gamma=0.95, c=4000$) & $\mathbf{72.79}\,{\scriptscriptstyle\pm0.45}$ & $7.08{\scriptstyle\pm0.40}$ & $0.23{\scriptstyle\pm0.00}$ & $7.08{\scriptstyle\pm0.40}$\\

    \bottomrule
  \end{tabular}
\end{table}

\begin{figure}[ht]
  \centering
  \begin{subfigure}[b]{0.24\columnwidth}
    \centering
    \includegraphics[width=\linewidth]{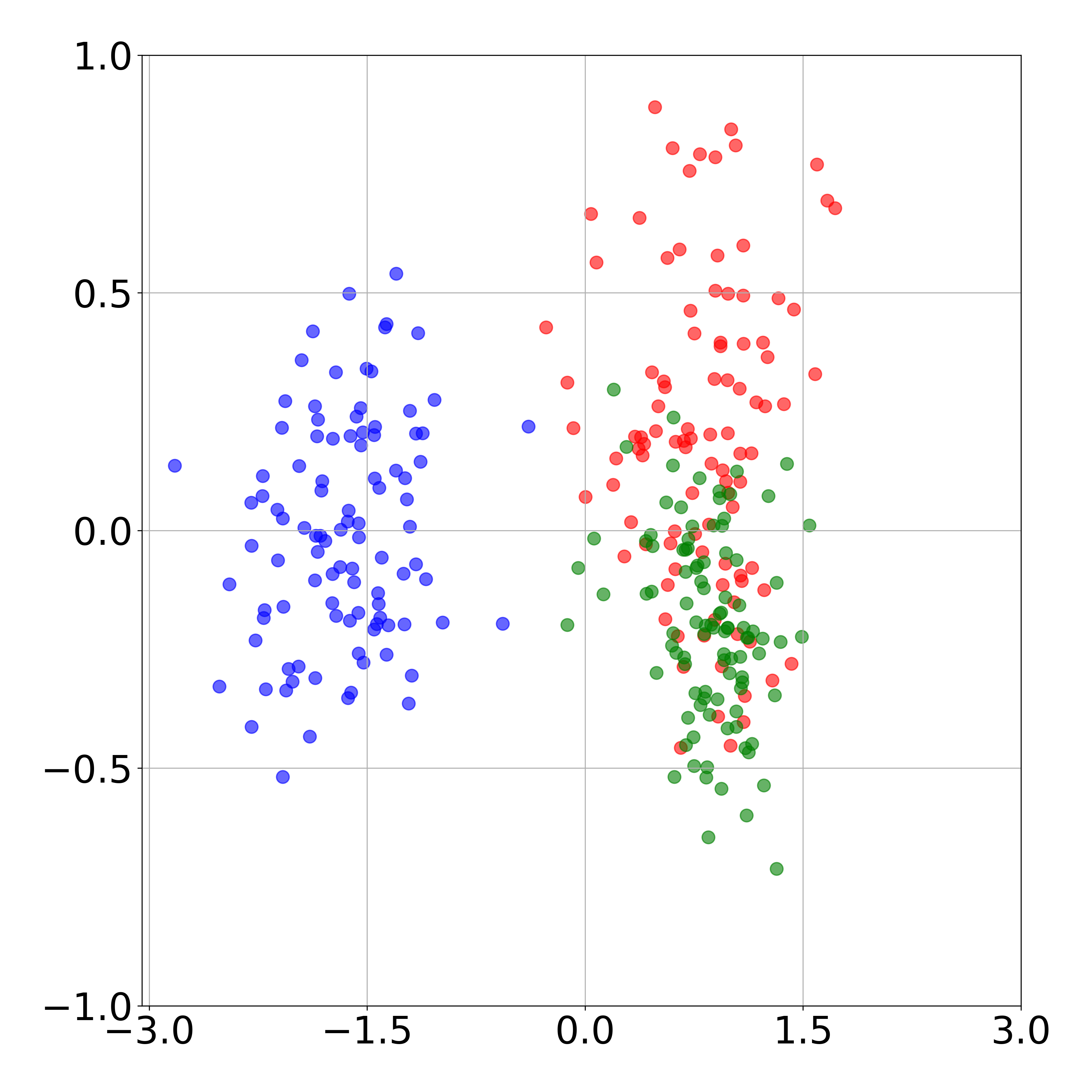}
    \caption{Baseline}
  \end{subfigure}
  \begin{subfigure}[b]{0.24\columnwidth}
    \centering
    \includegraphics[width=\linewidth]{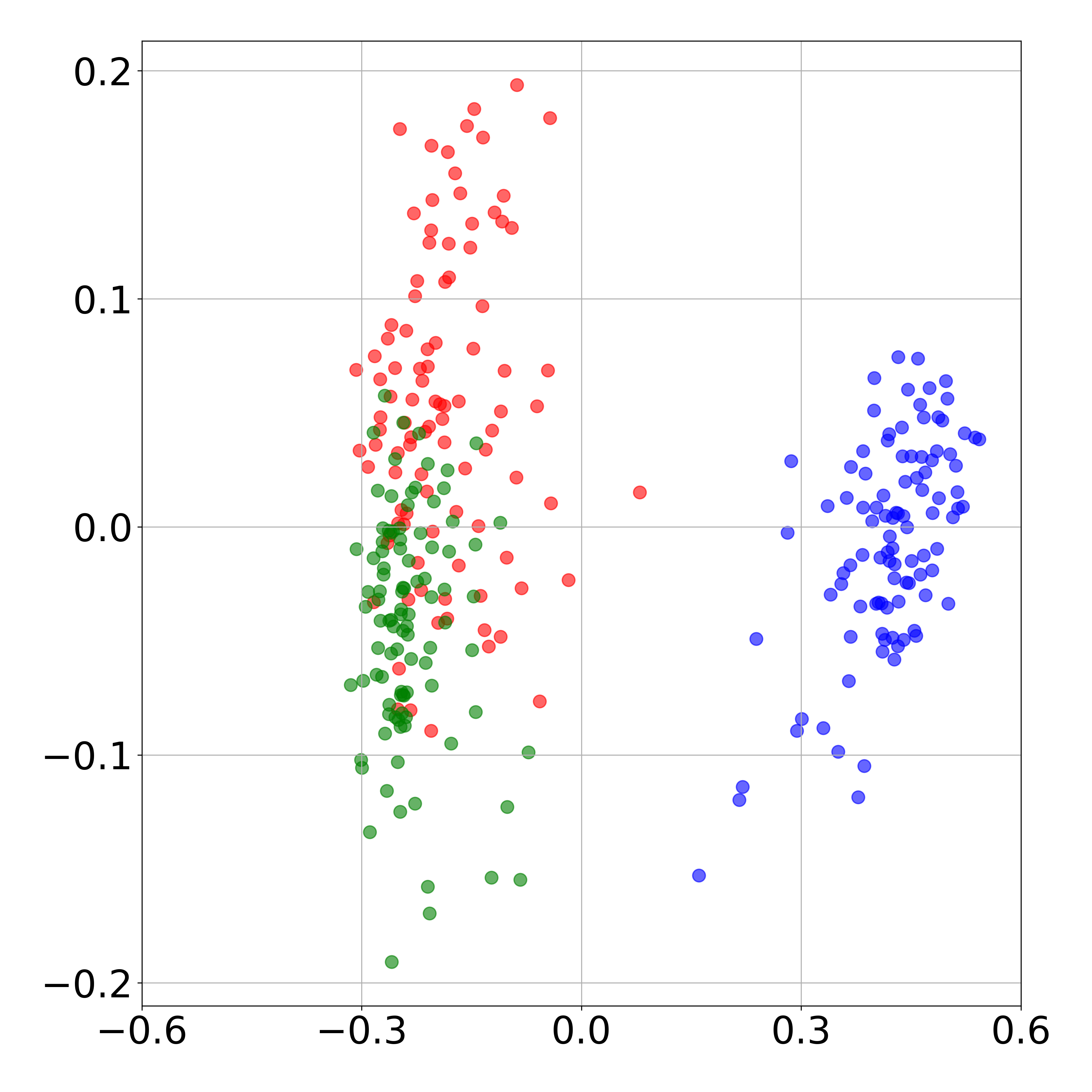}
    \caption{DPS}
  \end{subfigure}
    \begin{subfigure}[b]{0.24\columnwidth}
    \centering
    \includegraphics[width=\linewidth]{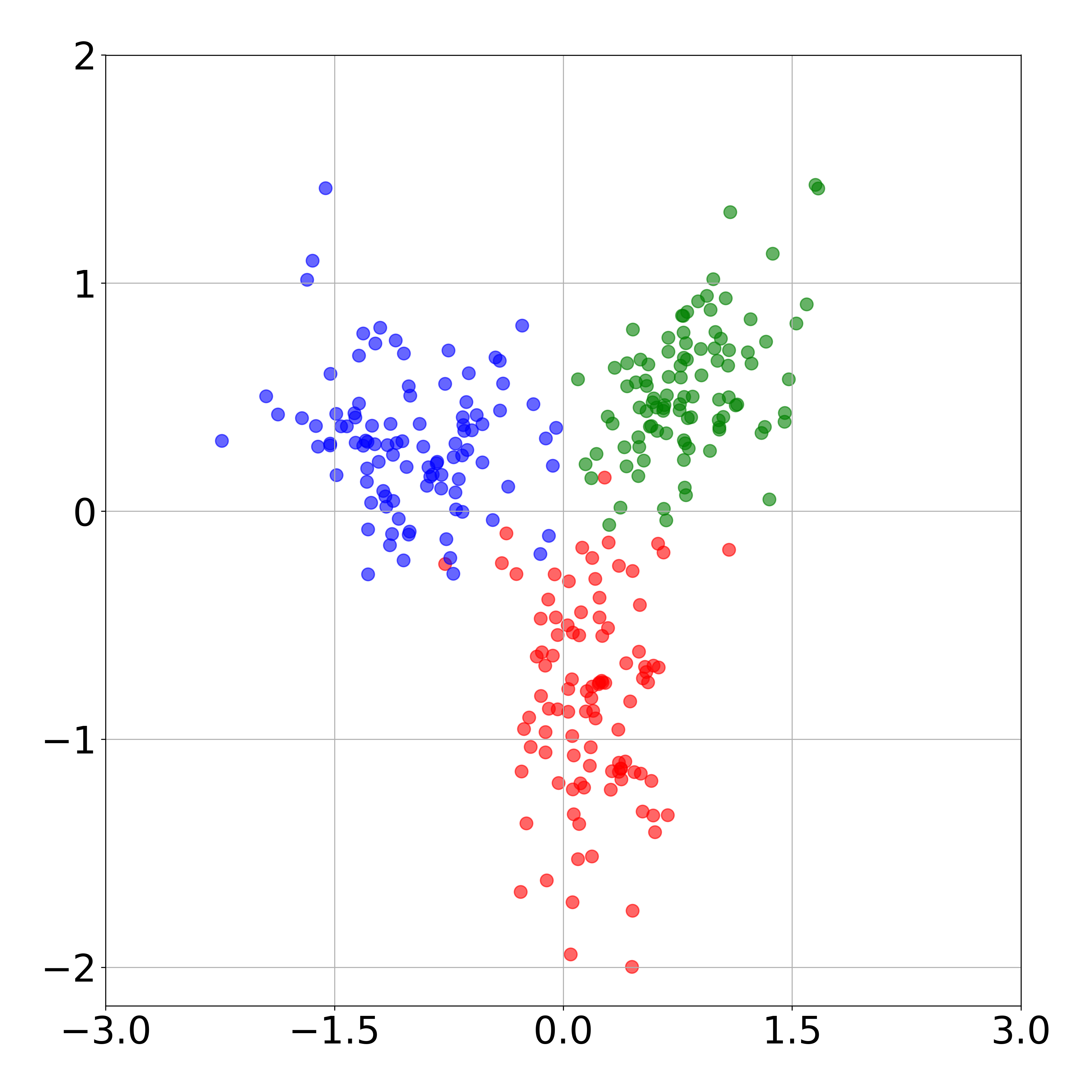}
    \caption{Baseline}
  \end{subfigure}
  \begin{subfigure}[b]{0.24\columnwidth}
    \centering
    \includegraphics[width=\linewidth]{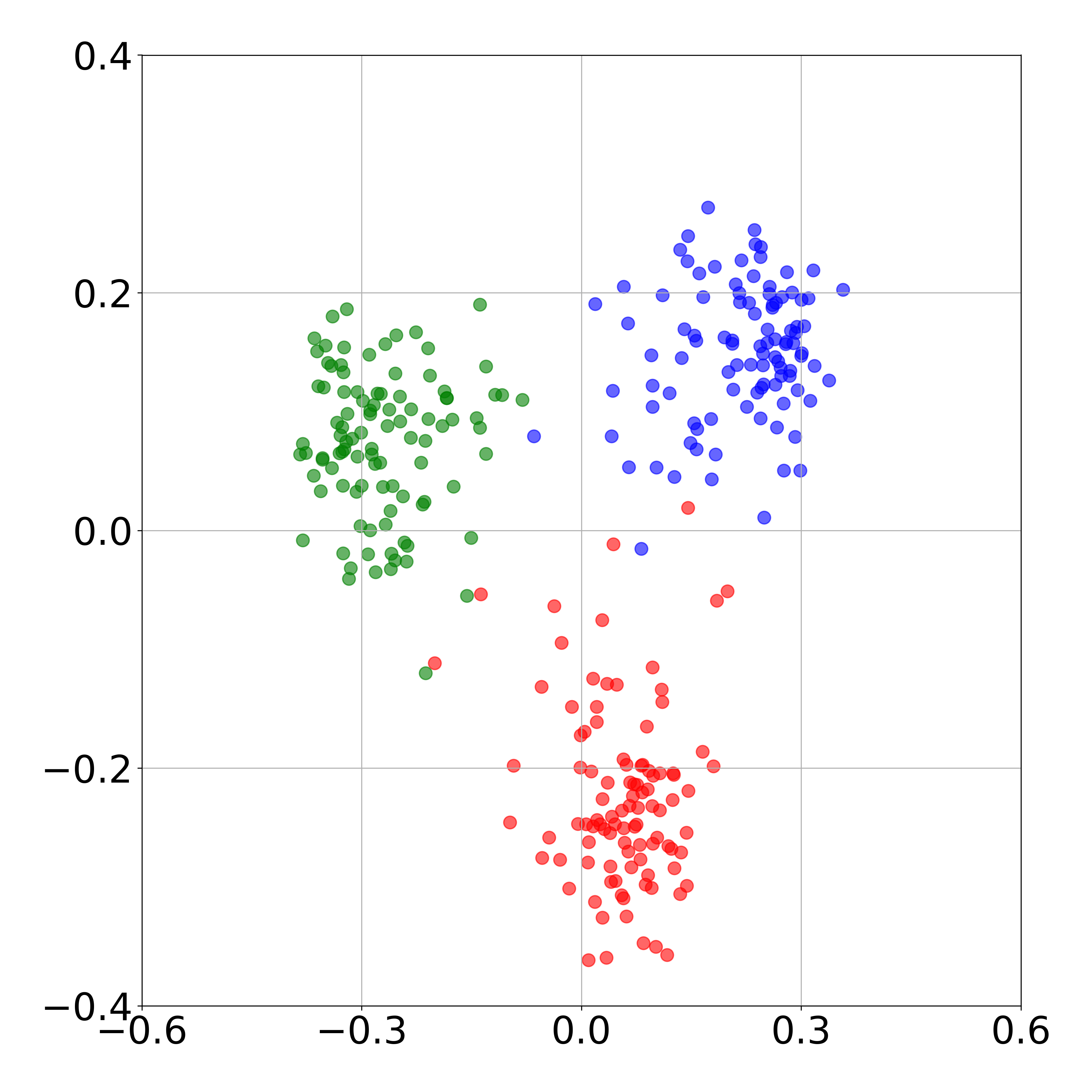}
    \caption{DPS}
  \end{subfigure}
    \caption{\textbf{Penultimate layer representations of ResNet-50 on CIFAR-100.} (a,b): Two semantically similar classes with one dissimilar class. (c,d): Three semantically dissimilar classes.}
  \label{fig:penultimate_representations}
\end{figure}

To study the emergence of dark knowledge during training, we compute the average KL divergence between the output distributions of the model over epochs and those of the final model, while adjusting for temperature scaling \citep{guo2017calibration}. The results are plotted in Figure \ref{fig:dk}, where the decrease of KL divergence over epochs suggest that dark knowledge is a property that emerges gradually.

\textbf{Out-of-distribution calibration and detection.} To evaluate model calibration under distributional shifts, we tested models trained on CIFAR-10 against the CIFAR-10-C benchmark, which applies 19 different corruptions (e.g., brightness, blur, noise) across five severity levels. As shown in Figure \ref{fig:calibration_cifar10c}, DPS consistently achieves the lowest Expected Calibration Error (ECE) across all severity levels compared to the baseline and related methods. Importantly, the performance gap widens as data quality degrades, and we observe that the related methods' ECE increases more rapidly at high corruption severities, while DPS maintains a flatter ECE curve. This indicates that DPS reduces overconfidence under increasing covariate shift.

Furthermore, we study the performance of DPS and related methods under domain shifts by measuring the area under the receiver operating characteristic curve (AUROC) for models trained on CIFAR-10 and CIFAR-100, evaluated against the Street View House Numbers (SVHN) dataset \cite{netzer2011reading}. While the best performing method varies with dataset and model architecture, we observe that DPS yields the most consistent improvement across datasets and architectures. Notably, while TE exhibits instability under this shift on CIFAR-100 (e.g., degrading performance on DenseNet-169 trained on CIFAR-100 with respect to baseline), DPS maintains robust performance. For ViT-B, we observe performance saturation on CIFAR-10 (with all methods $>98\%$), while for CIFAR-100, DPS yields a notable improvement ($+2.29\%$) over the baseline.

\textbf{Calibration.} We train a WideResNet-40-1 on CIFAR-100 to benchmark DPS against different calibration methods including distillation-based method. We compare against Label Smoothing (LS) \citep{szegedy2016rethinking}, Temperature Scaling (TS) \citep{guo2017calibration}, MixUp \citep{zhang2017mixup}, Correctness Ranking
Loss (CRL) \citep{moon2020confidence}, PS-KD \citep{kim2021self}, Multi-class Difference
in Confidence and Accuracy (MDCA) \citep{hebbalaguppe2022stitch}, AdaFocal \citep{ghosh2022adafocal},  Calibration by Pairwise Constraints (CPC) \citep{cheng2022calibrating},  Margin-based Label Smoothing (MbLS) \citep{liu2022devil}, Adaptive and Conditional Label Smoothing (ACLS) \citep{park2023acls} and combinations of the aforementioned method with knowledge distillation \citep{hebbalaguppe2024calibration}. The results are included in Table \ref{tab:calibration_wrn}, where we observe that DPS yields the highest accuracy and the lowest ECE, Static Calibration Error (SCE) and Adaptive Calibration Error (ACE) of all methods.

\subsection{Penultimate Layer Representations}
Inspired by \cite{muller2019does}, we visualize the penultimate layer representations in Figure \ref{fig:penultimate_representations}. DPS yields tighter, less overlapping clusters than conventional training, which is somewhat surprising since DPS promotes learning of sample-specific features. It seems that learning similarities between classes can help differentiate among them.

\subsection{Limitations}
While we experiment with various forms of data augmentation, the interaction with different augmentation schemes as well as regularization techniques warrants further study. Intuitively, augmentations that increase prediction variance may benefit from higher values of the discount factor $\gamma$. Finally, DPS requires selecting the discount factor $\gamma$ and the prior strength $c$, which, despite the observed performance across a wide range of settings, could be viewed as a methodological limitation.

\end{document}